\documentclass[10pt,twocolumn,letterpaper]{article}

\usepackage[pagenumbers]{iccv} %

\usepackage[T1]{fontenc}
\usepackage{tikz,subcaption}
\usetikzlibrary{spy,backgrounds,positioning,shapes,arrows}
\usepackage{xcolor}
\usepackage{amsmath}
\usepackage[ruled]{algorithm2e}
\usepackage{algpseudocode}
\usepackage{nicefrac}
\usepackage{indentfirst}
\usepackage{tabularx}
\usepackage{color}
\usepackage{colortbl}
\usepackage{graphicx}
\usepackage{lipsum}
\usepackage{wrapfig}
\usepackage{graphicx}
\usepackage{caption}
\usepackage{subcaption}
\usepackage{tikz}
\usepackage{pgfplots}
\usetikzlibrary{spy,calc}
\newif\ifblackandwhitecycle
\gdef\patternnumber{0}
\pgfkeys{/tikz/.cd,
    zoombox paths/.style={
        draw=orange,
        very thick
    },
    black and white/.is choice,
    black and white/.default=static,
    black and white/static/.style={ 
        draw=white,   
        zoombox paths/.append style={
            draw=white,
            postaction={
                draw=black,
                loosely dashed
            }
        }
    },
    black and white/static/.code={
        \gdef\patternnumber{1}
    },
    black and white/cycle/.code={
        \blackandwhitecycletrue
        \gdef\patternnumber{1}
    },
    black and white pattern/.is choice,
    black and white pattern/0/.style={},
    black and white pattern/1/.style={    
            draw=white,
            postaction={
                draw=black,
                dash pattern=on 2pt off 2pt
            }
    },
    black and white pattern/2/.style={    
            draw=white,
            postaction={
                draw=black,
                dash pattern=on 4pt off 4pt
            }
    },
    black and white pattern/3/.style={    
            draw=white,
            postaction={
                draw=black,
                dash pattern=on 4pt off 4pt on 1pt off 4pt
            }
    },
    black and white pattern/4/.style={    
            draw=white,
            postaction={
                draw=black,
                dash pattern=on 4pt off 2pt on 2 pt off 2pt on 2 pt off 2pt
            }
    },
    zoomboxarray inner gap/.initial=5pt,
    zoomboxarray columns/.initial=2,
    zoomboxarray rows/.initial=2,
    subfigurename/.initial={},
    figurename/.initial={zoombox},
    zoomboxarray/.style={
        execute at begin picture={
            \begin{scope}[
                spy using outlines={%
                    zoombox paths,
                    width=\imagewidth / \pgfkeysvalueof{/tikz/zoomboxarray columns} - (\pgfkeysvalueof{/tikz/zoomboxarray columns} - 1) / \pgfkeysvalueof{/tikz/zoomboxarray columns} * \pgfkeysvalueof{/tikz/zoomboxarray inner gap} -\pgflinewidth,
                    height=\imageheight / \pgfkeysvalueof{/tikz/zoomboxarray rows} - (\pgfkeysvalueof{/tikz/zoomboxarray rows} - 1) / \pgfkeysvalueof{/tikz/zoomboxarray rows} * \pgfkeysvalueof{/tikz/zoomboxarray inner gap}-\pgflinewidth,
                    magnification=3,
                    every spy on node/.style={
                        zoombox paths
                    },
                    every spy in node/.style={
                        zoombox paths
                    }
                }
            ]
        },
        execute at end picture={
            \end{scope}
            \node at (image.north) [anchor=north,inner sep=0pt] {\subcaptionbox{\label{\pgfkeysvalueof{/tikz/figurename}-image}}{\phantomimage}};
            \node at (zoomboxes container.north) [anchor=north,inner sep=0pt] {\subcaptionbox{\label{\pgfkeysvalueof{/tikz/figurename}-zoom}}{\phantomimage}};
     \gdef\patternnumber{0}
        },
        spymargin/.initial=0.5em,
        zoomboxes xshift/.initial=1,
        zoomboxes right/.code=\pgfkeys{/tikz/zoomboxes xshift=1},
        zoomboxes left/.code=\pgfkeys{/tikz/zoomboxes xshift=-1},
        zoomboxes yshift/.initial=0,
        zoomboxes above/.code={
            \pgfkeys{/tikz/zoomboxes yshift=1},
            \pgfkeys{/tikz/zoomboxes xshift=0}
        },
        zoomboxes below/.code={
            \pgfkeys{/tikz/zoomboxes yshift=-1},
            \pgfkeys{/tikz/zoomboxes xshift=0}
        },
        caption margin/.initial=4ex,
    },
    adjust caption spacing/.code={},
    image container/.style={
        inner sep=0pt,
        at=(image.north),
        anchor=north,
        adjust caption spacing
    },
    zoomboxes container/.style={
        inner sep=0pt,
        at=(image.north),
        anchor=north,
        name=zoomboxes container,
        xshift=\pgfkeysvalueof{/tikz/zoomboxes xshift}*(\imagewidth+\pgfkeysvalueof{/tikz/spymargin}),
        yshift=\pgfkeysvalueof{/tikz/zoomboxes yshift}*(\imageheight+\pgfkeysvalueof{/tikz/spymargin}+\pgfkeysvalueof{/tikz/caption margin}),
        adjust caption spacing,
    },
    calculate dimensions/.code={
        \pgfpointdiff{\pgfpointanchor{image}{south west} }{\pgfpointanchor{image}{north east} }
        \pgfgetlastxy{\imagewidth}{\imageheight}
        \global\let\imagewidth=\imagewidth
        \global\let\imageheight=\imageheight
        \gdef\columncount{1}
        \gdef\rowcount{1}
        
    },
    image node/.style={
        inner sep=0pt,
        name=image,
        anchor=south west,
        append after command={
            [calculate dimensions]
            node [image container,subfigurename=\pgfkeysvalueof{/tikz/figurename}-image] {\phantomimage}
            node [zoomboxes container,subfigurename=\pgfkeysvalueof{/tikz/figurename}-zoom] {\phantomimage}
        }
    },
    color code/.style={
        zoombox paths/.append style={draw=#1}
    },
    connect zoomboxes/.style={
    spy connection path={\draw[draw=none,zoombox paths] (tikzspyonnode) -- (tikzspyinnode);}
    },
    help grid code/.code={
        \begin{scope}[
                x={(image.south east)},
                y={(image.north west)},
                font=\footnotesize,
                help lines,
                overlay
            ]
            \foreach \x in {0,1,...,9} { 
                \draw(\x/10,0) -- (\x/10,1);
                \node [anchor=north] at (\x/10,0) {0.\x};
            }
            \foreach \y in {0,1,...,9} {
                \draw(0,\y/10) -- (1,\y/10);                        \node [anchor=east] at (0,\y/10) {0.\y};
            }
        \end{scope}    
    },
    help grid/.style={
        append after command={
            [help grid code]
        }
    },
}
\newcommand\phantomimage{%
    \phantom{%
        \rule{\imagewidth}{\imageheight}%
    }%
}
\newcommand\zoombox[2][]{
    \begin{scope}[zoombox paths]
        \pgfmathsetmacro\xpos{
            (\columncount-1)*(\imagewidth / \pgfkeysvalueof{/tikz/zoomboxarray columns} + \pgfkeysvalueof{/tikz/zoomboxarray inner gap} / \pgfkeysvalueof{/tikz/zoomboxarray columns} ) + \pgflinewidth
        }
        \pgfmathsetmacro\ypos{
            (\rowcount-1)*( \imageheight / \pgfkeysvalueof{/tikz/zoomboxarray rows} + \pgfkeysvalueof{/tikz/zoomboxarray inner gap} / \pgfkeysvalueof{/tikz/zoomboxarray rows} ) + 0.5*\pgflinewidth
        }
        \edef\dospy{\noexpand\spy [
            #1,
            zoombox paths/.append style={
                black and white pattern=\patternnumber
            },
            every spy on node/.append style={#1},
            x=\imagewidth,
            y=\imageheight
        ] on (#2) in node [anchor=north west] at ($(zoomboxes container.north west)+(\xpos pt,-\ypos pt)$);}
        \dospy
        \pgfmathtruncatemacro\pgfmathresult{ifthenelse(\columncount==\pgfkeysvalueof{/tikz/zoomboxarray columns},\rowcount+1,\rowcount)}
        \global\let\rowcount=\pgfmathresult
        \pgfmathtruncatemacro\pgfmathresult{ifthenelse(\columncount==\pgfkeysvalueof{/tikz/zoomboxarray columns},1,\columncount+1)}
        \global\let\columncount=\pgfmathresult
        \ifblackandwhitecycle
            \pgfmathtruncatemacro{\newpatternnumber}{\patternnumber+1}
            \global\edef\patternnumber{\newpatternnumber}
        \fi
    \end{scope}
}

\usepackage{multirow}
\usepackage{textcomp}
\usepackage{gensymb} %
\usepackage{bm} %
\usepackage{lmodern}
\pgfplotsset{compat=1.18} %

\allowdisplaybreaks %

\def\bx{{\bf x}}
\def\by{{\bf y}}
\def\bxa{{\bf x}^{acq}_i}
\def\bya{{\bf y}^{acq}_i}
\def\bw{{\bf w}}
\def\bc{{\bf c}}
\def\bH{{\mathbf H}}
\DeclareMathOperator{\tr}{tr}

\DeclareMathOperator*{\argmin}{arg\,min}
\newcommand{\Cov}{\mathrm{Cov}}

\DeclareMathOperator{\diag}{diag}

\newcommand{\meanC}[0]{\boldsymbol{\mu}_{C}}
\newcommand{\meanI}[0]{\boldsymbol{\mu}_{I}}

\newcommand{\covI}[0]{\boldsymbol{\Sigma}_{I}}
\newcommand{\cam}[0]{\bx}
\newcommand{\camCW}[0]{\bx}

\newcommand{\tauC}[0]{\tau}

\newcommand{\Exp}[0]{\text{Exp}}
\newcommand{\Log}[0]{\text{Log}}

\newcommand{\pd}[2]{\frac{\partial {#1} }{\partial {#2} }}
\newcommand{\mpd}[2]{\frac{\mathcal{D} {#1}}{\mathcal{D} {#2}}}
\newcommand{\se}[1]{\mathfrak{se}(#1)}

\newcommand{\identity}[0]{\boldsymbol{I}}

\newcommand{\matW}[0]{\mathbf{W}}

\usepackage{rotating}
\usepackage{listings}
\usepackage{courier}

\makeatletter
\renewcommand{\paragraph}{%
  \@startsection{paragraph}{4}%
  {\z@}{2.0ex \@plus 1ex \@minus .2ex}{-1em}%
  {\normalfont\normalsize\bfseries}%
}
\makeatother

\definecolor{iccvblue}{rgb}{0.21,0.49,0.74}
\usepackage[pagebackref,breaklinks,colorlinks,allcolors=iccvblue]{hyperref}

\def\paperID{6753} %
\def\confName{ICCV}
\def\confYear{2025}

\title{Multimodal LLM Guided Exploration and Active Mapping using Fisher Information}

\author{Wen Jiang\footnotetext{* equal contribution}\textsuperscript{*}\textsuperscript{1} \and Boshu Lei\textsuperscript{*}\textsuperscript{1} \and Katrina Ashton\textsuperscript{1} \and Kostas Daniilidis\textsuperscript{1,2}
\\
\textsuperscript{1}~University of Pennsylvania \quad \textsuperscript{1,2}~Archimedes, Athena RC
}

\begin{document}
\maketitle
\begin{abstract}
We present an active mapping system which plans for both long-horizon exploration goals and short-term actions using a 3D Gaussian Splatting (3DGS) representation. 
Existing methods either do not take advantage of recent developments in multimodal Large Language Models (LLM) or do not consider challenges in localization uncertainty, which is critical in embodied agents.
We propose employing multimodal LLMs for long-horizon planning in conjunction with detailed motion planning using our information-based objective.
By leveraging high-quality view synthesis from our 3DGS representation, our method employs a multimodal LLM as a zero-shot planner for long-horizon exploration goals from the semantic perspective. 
We also introduce an uncertainty-aware path proposal and selection algorithm that balances the dual objectives of maximizing the information gain for the environment while minimizing the cost of localization errors. 
Experiments conducted on the Gibson and Habitat-Matterport 3D datasets demonstrate state-of-the-art results of the proposed method.
\end{abstract}

\section{Introduction}
Being able to autonomously explore and map an environment while localizing within that map is a core skill for a mobile robot.
This ability could empower embodied artificial intelligence systems with effective 3D scene understanding through use in conjunction with vision-language features~\cite{zhou2024feature, shi2024language, qin2024langsplat, yulanguage} and can be used as a basis for language-specified robotics tasks~\cite{hou2024tamma,shafiullah2023clipfields, liu2024okrobot, huang2023visual}.
This task is challenging because it requires general knowledge of the typical layout of an environment to identify coarse targets for long-term exploration, as well as the ability to plan each step to ensure localization accuracy while maximizing information gain or `surprise' when traversing the environment. 

\begin{figure}[t]
    \centering
\includegraphics[width=\linewidth]{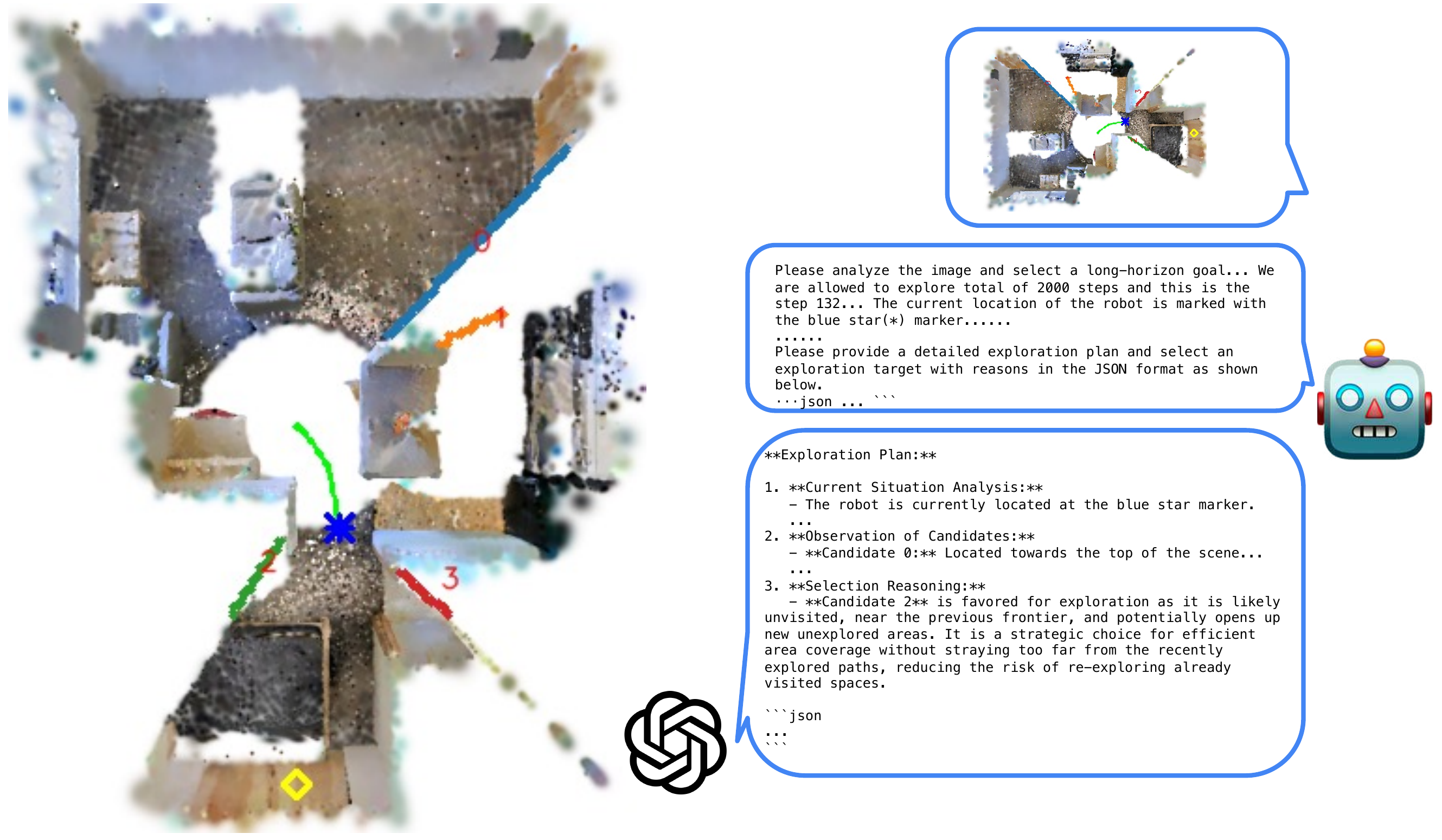}
    \caption{{\bf An overview about the interaction with a multimodal LLM.} The image on the left is a bird's-eye view rendering from our 3DGS representation. A detailed example of the dialog can be found in the supplementary.}
    \label{fig:chat}
\end{figure}

We use 3D Gaussian Splatting (3DGS)~\cite{kerbl3dgaussians} as scene representation and tackle this problem in two phases: long-horizon planning with a Large Language Model (LLM), and detailed motion planning with our information-based approach by quantifying the Fisher Information for 3DGS parameters and current localization states. 
As shown in \cref{fig:chat}, we prompt a multimodal LLM with our current map state, trajectory, and frontiers~\cite{yamauchi1997frontier}, used as candidates for long-term exploration to exploit the LLM's knowledge about scene layouts. We then follow the suggestion from the LLM by choosing paths that maximize the Expected Information Gain (EIG) calculated using the Fisher Information about both poses and the scene. 

Classical active mapping methods often define objectives to reduce the uncertainty of the state estimate based on its covariance~\cite{feder1999adaptive,rodriguez2018importance,carrillo2012comparison,vidal2006active,chen2020active,Semantic-OcTree}. The covariance is readily available for classical systems which use filters to update the state estimate. 
However, many recent mapping systems use non-linear optimization to update the state estimate, making the covariance difficult to obtain. In particular, systems with 3DGS for the scene representation~\cite{matsuki2023gaussian,yan2023gs,keetha2023splatam,hhuang2024photoslam} have been developed for high-fidelity rendering of novel views of the scene.
Previous methods attempted to quantify the uncertainty of radiance fields for reconstructing scenes from given data~\cite{shen2021snerf,cf-nerf, sunderhauf2023density, goli2023} for active view selection~\cite{pan2022activenerf,jiang2024fisherrf, sunderhauf2023density, lee2022uncertainty} and active object reconstruction~\cite{yan2023activeIO} or mapping~\cite{yan2023active-neural-mapping,kuang2024active,jin2024gs,jiang2024fisherrf} of scenes with given localization, and for 3D reconstruction of small scenes with an inward-facing camera~\cite{zhan2022activermap}. 
However, all the prior methods only model the uncertainty of the scene representation, ignoring another primary source of error for a practical mapping system: the risk of localization failures.
Our proposed system could address not only Information Gain during exploration but also consider localization accuracy when the robot explores unknown and texture-less regions.

In terms of long-horizon planning for robot exploration, classical approaches such as frontier-based exploration~\cite{yamauchi1997frontier} and A* algorithms are still used in active mapping systems for their efficiency and simplicity~\cite{stachniss2004exploration,carlone2014active,trivun2015active,sun2020frontier,kim2015active}. 
However, algorithms that use only simple heuristics cannot determine the information gain, nor estimate 3D geometry, limiting their objective to simply improving coverage or having minimal travel distances. 
To address this issue, learnable approaches have been used for exploration tasks~\cite{upen,informative-uav,liu2023active,morilla2023robust}, but these methods are trained for a limited scene distribution. %
Recently, LLMs and Vision Language Models have demonstrated extraordinary abilities in visual grounding and logical reasoning and have been studied with various robotic tasks~\cite{sermanet2024robovqa,huang2023visual,team2024octo,chen2024affordances,chen2024mapgpt,du2023video,driess2023palme}.
However, these works do not focus on pure exploration or reconstruction tasks and many do not take advantage of photo-realistic 3D representations for multimodal LLMs.

\begin{figure}[t]
    \centering
    \includegraphics[width=\linewidth]{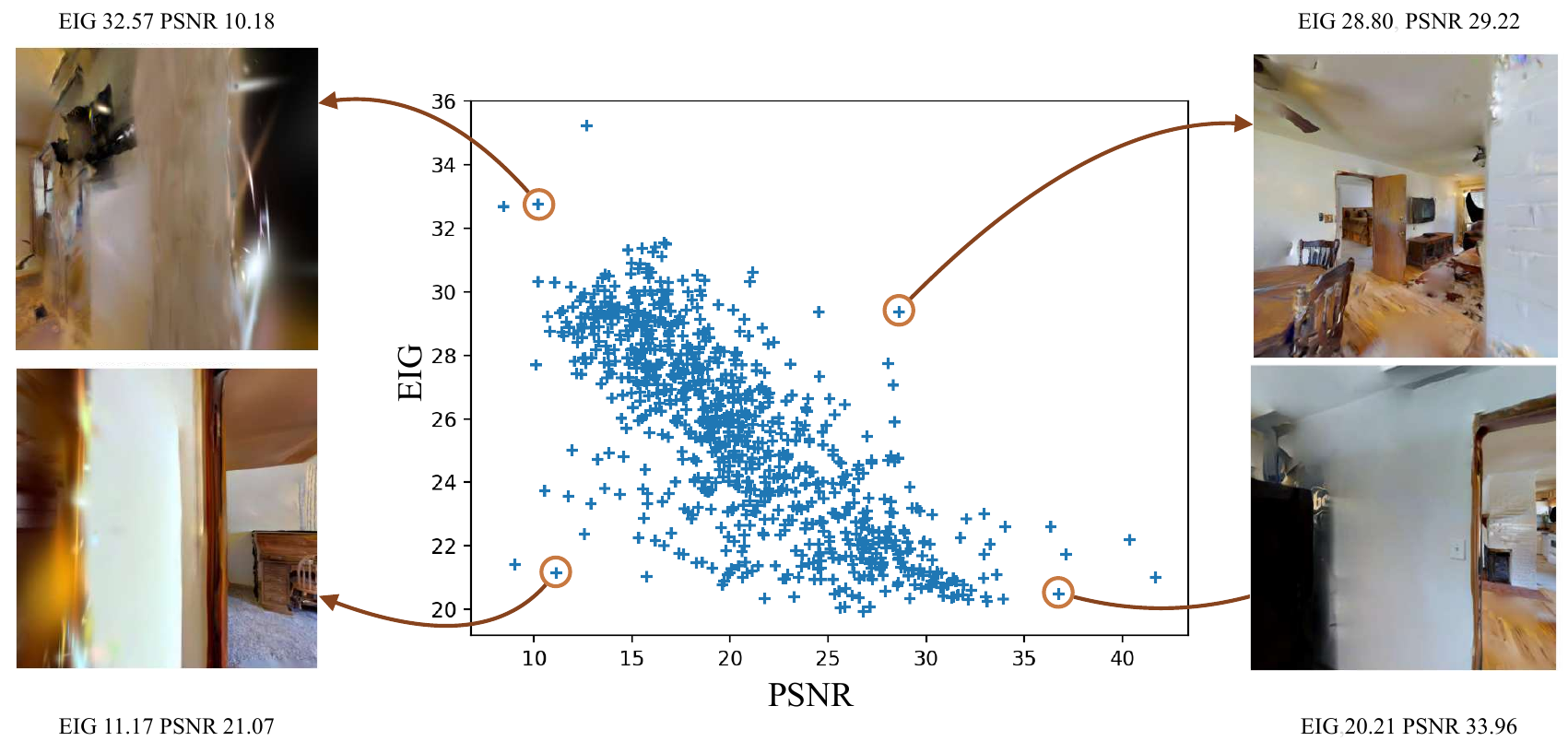}
    \caption{\textbf{ Scatter Plot of EIG vs. PSNR } We plot the EIG and PSNR at sampled poses in the Cantwell scene of the Gibson dataset. The figure corroborates the intuition that the robot expects to gain little information (low EIG) at well reconstructed region (high PSNR) and gain much information (high EIG) at a poorly reconstructed region (low PSNR).  }
    \label{fig:eig-vs-psnr}
\end{figure}
We argue that we leverage the best of both worlds by using a multi-modal LLM to identify long-horizon exploration goals and planning paths to them using our information-based algorithm. Our approach is seamlessly integrated with our 3DGS scene representation, we utilize a high-quality bird's-eye view rendering from the scene representation to prompt the LLM and can directly extract the EIG for path planning from it using Fisher Information.  
As shown in Fig.~\ref{fig:eig-vs-psnr}, our EIG metric reflects the `surprise' and correlates well with the rendering quality on candidate views, as measured by the peak signal-to-noise ratio (PSNR) between the true and rendered image, without actually taking a picture at the candidate location.
Notably, we compute the full Fisher Information matrix for localization parameters without any approximation with our efficient CUDA implementation.
We also derived localization uncertainty on candidate paths with the Cramér–Rao bound, and we use this along with the EIG in our final objective to balance the dual problem of exploration and localization. 
To validate our approach, we evaluate our method on scenes from the Gibson~\cite{xiazamirhe2018gibsonenv} and Habitat-Matterport 3D~\cite{ramakrishnan2021hm3d} datasets quantitatively and qualitatively. 
We show superior reconstruction quality in various metrics compared to several baselines and recent state-of-the-art methods~\cite{chaplot2020learning,placed2022explorb,upen,yan2023active-neural-mapping,yamauchi1997frontier}.

\noindent
Our contributions can be summarized as follows:
\begin{itemize}
    \item We present an active mapping system for ground robots that could autonomously explore the environment and extensively compare our system with previous approaches. To the best of our knowledge, we are the first active mapping system with 3D Gaussian representation that is not dependent on ground truth camera pose readings. 
    \item We provide a way to leverage the zero-shot long-horizon planning ability of LLM into our active mapping system seamlessly.
    \item We introduce localization uncertainty with active mapping systems and effectively balance the information gain for exploration and the cost of possible localization errors. 

\end{itemize}

\section{Related work}
\paragraph{Active Mapping and Localization}
Efficiently exploring an environment in order to map it while being able to localize in that map is a fundamental problem in robotics. There are many methods that address aspects of this problem separately -- Simultaneous Localization and Mapping (SLAM) methods~\cite{mur2015orb,engel2017direct,mur2017orb,campos2021orb,izadi2011kinectfusion,newcombe2011dtam,engel2014lsd,teed2021droid, tao20243d} address the mapping and localization. Many exploration methods~\cite{upen,chen2018learning,yan2023active-neural-mapping,kuang2024active,jin2024gs,jiang2024fisherrf} address the exploration and mapping aspects while assuming poses are provided. Active SLAM methods~\cite{stachniss2004exploration,carlone2014active,trivun2015active,sun2020frontier} consider both of these problems. The exploration in these systems is usually driven by a measure of uncertainty~\cite{bry2011rapidly,he2010puma,lluvia2021active-survey,placed2023survey}; specific utility functions are often drawn from either Information Theory (IT)~\cite{shannon1948mathematical}, or the Theory of Optimal Experimental Design (TOED)~\cite{pazman1986foundations}. Recently, learning-based approaches have been developed for active mapping and localization; Active Neural SLAM~\cite{chaplot2020learning} learns policies to drive exploration and estimate the agent pose, more similarly to us NARUTO~\cite{feng2024naruto} uses an SLAM backbone to estimate the pose and uses uncertainty to drive exploration. However, they only consider the reconstruction uncertainty, whereas we choose long-horizon planning goals and consider the localization uncertainty as well as the reconstruction uncertainty when planning the best path to the long-horizon goal.

\paragraph{Uncertainty quantification for radiance fields}
The vast majority of previous work on uncertainty quantification for radiance fields has been for post-processing scenes~\cite{shen2021snerf,cf-nerf, sunderhauf2023density, goli2023, tao2024rt}, view selection~\cite{pan2022activenerf,jiang2024fisherrf, sunderhauf2023density, lee2022uncertainty} or active view selection~\cite{pan2022activenerf,sunderhauf2023density,jiang2024fisherrf}, all of which assume the input images are posed. 
Active neural mapping~\cite{yan2023active-neural-mapping,kuang2024active} uses neural variability, that is, the prediction robustness against random weight perturbations, as an estimate of uncertainty to actively map a scene with ground truth poses provided. Fisher-RF~\cite{jiang2024fisherrf} also performs active scene mapping with ground truth poses provided, based on an approximation of the Fisher Information of views along candidate paths. \citet{zhan2022activermap} performs active reconstruction without ground truth camera poses. However, they only evaluate small-scale scenes and limit the camera trajectories to be inwards facing and only model scene uncertainty, not localization uncertainty, which is a key consideration for active mapping and exploration. 

\paragraph{Robot Planning with Foundation Models}
LLMs and VLMs have been used widely in robotics as for language-specified tasks, either as high-level planners in conjunction with other methods for low-level control~\cite{huang2022language,huang2023visual,long2024instructnav,song2023llm,brohan2023can,rana2023sayplan,du2023video,driess2023palme}, to create intermediate representations which can be planned over~\cite{huang2023voxposer} or directly outputting actions~\cite{Zhou_Hong_Wu_2024,long2024discuss}. 
Outside of language-driven tasks, VLMs and LLMs have been used in robotics for providing rewards to drive exploration in Reinforcement Learning~\cite{triantafyllidis2024intrinsic}, and for visual localization~\cite{mirjalili2023fmloc}.
More similar to our task, LLMs and VLMs have also been used for goal-driven navigation, such as searching for specific objects~\cite{yokoyama2024vlfm,s2024cognitive,dorbala2023can,shah2023navigation}.

\section{Method}

\begin{figure*}[t]
    \centering
\includegraphics[width=\textwidth]{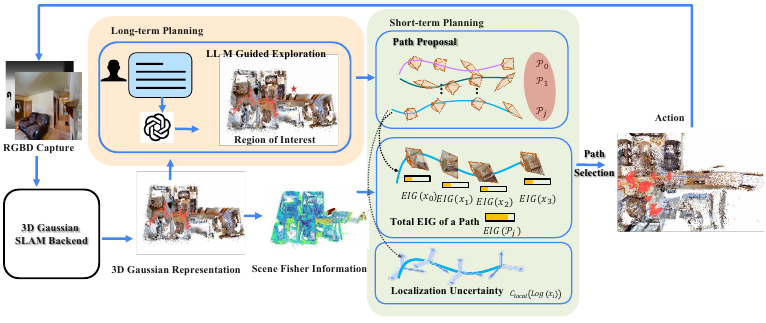}
    \caption{{\bf An Illustration of Our Active Mapping System} Our method first identifies long-horizon targets with a multimodal LLM by using novel-view synthesis from our 3DGS representation.
    Our information-based path proposal method then proposes and selects short-term action sequences from the region of interest the multimodal LLM identifies. The best path and action along the path is selected concerning both the information gain and localization accuracy.}
    \label{fig:agslam-pipeline}
\end{figure*}

We divide the scene exploration task into two phases: (a) long-horizon planning for the coarse direction of exploration that leads to better coverage and understanding (Sec.~\ref{sec:long-term}), and (b) detailed trajectory proposal and path selection that aims to improve 3D reconstruction (Sec.~\ref{sec: path eig}) and localization (Sec.~\ref{ssec:path localization}) from a geometrical perspective. 
The former task relies more on semantic information and prior knowledge about the possible layout of the environment, which is suitable for a generalist model with prior distributions on the scene.
The latter task, however, involves detailed motion planning and is better addressed by considering information gain on fine-level geometries. \cref{fig:agslam-pipeline} shows an overview of our method.

\subsection{Preliminary}\label{sec:pre}

In 3D Gaussian Splatting (3DGS)~\cite{kerbl3dgaussians}, the scene is represented by a set of 3D Gaussians whose color and opacity are learned via a rendering loss. An image can be rendered by projecting the Gaussians to 2D and using $\alpha$-blending for the $N$ ordered points on the 2D splat that overlaps each pixel. The Jacobian of the current camera pose $\bx$ with respect to the parameters of 3D Gaussians can be computed by defining the (left) partial derivative on the manifold~\cite{matsuki2023gaussian}:
\begin{equation}
    \mpd{f(\cam)}{\cam} \triangleq  \lim_{\tauC \to 0}\frac{\Log(f(\Exp(\tau) \circ \cam) \circ f(\cam)^{-1})}{\tauC}~,
\end{equation}
where $\tauC \in \se{3}$, $\circ$ is a group composition, and $\Exp$ and $\Log$ are the exponential and logarithmic mappings between Lie algebra and Lie Group.

Fisher Information is a measurement of the information that a random variable $\by$ carries about an unknown parameter $\bw$ of a distribution that models $\by$. In the problem of novel view synthesis, 
we are interested in measuring the observed information of a radiance field with parameters $\bw$ at a camera pose $\bx$ using the negative log-likelihood of the image observation $\by$ taken from that pose:
\begin{equation}
-\log p(\by|\bx; \bw) = (\by - f(\bx, \bw))^T(\by - f(\bx, \bw)),
\label{eq:log-likelihood}
\end{equation}
where $f(\bx, \bw)$ is the rendering model. Under regularity conditions~\cite{schervish2012theory}, the Fisher Information of $-\log p(\by|\bx; \bw)$ is the Hessian of Eq.~\ref{eq:log-likelihood} with respect to $\bw$, denoted $\bH''[\by|\bx,\bw]$.
In our formulation $\by \in R^{\it{h}\times \it{w} \times \it{c}}$ is the RGB-D observation, and $\bw$ is a tensor for all the 3D Gaussian Parameters $R^{N\times 14}$, where $N$ is the number of Gaussians and the parameters are the 3D means, RGB colors, unormalized rotations represented as quaternions, opacities and per-axis log-scales.

FisherRF~\cite{jiang2024fisherrf} uses Fisher Information to address active perception problems such as active view selection and active mapping, although they do not consider localization uncertainty. Given a training set of views $D^{train}$ and a set of candidate trajectories $\{\mathcal{P}_j\}$ FisherRF selects the path which maximizes the following:
\begin{equation}
    \tr\left( \left(\sum_{\bxa \in \mathcal{P}_j} \bH''[\bya | \bxa, \bw] \right) \bH''[\bw| D^{train}]^{-1} \right)
    \label{eq:agslam-opt}
\end{equation}
where $\bw$ is the initial estimate of model parameters using the current training set. 
Crucially, the Fisher Information $\bH''[\bya | \bxa, \bw]$ does not depend on the label $\bya$ of the acquisition sample $\bxa$, thus it is feasible to compute it before visiting the potential view candidate $\bxa$. 
However, the number of optimizable parameters is typically more than 20 million, which means it is impossible to compute without sparsification or approximation. 
In practice, FisherRF~\cite{jiang2024fisherrf} applies a Laplace approximation~\cite{laplace2021, bayesian-interpolation} that approximates the Hessian matrix with its diagonal values plus a log-prior regularizer $\lambda I$ as follows
\begin{equation}\label{eq:agslam-hessian-reg}
    \bH''[\by|\bx, \bw] \simeq \diag(\nabla_\bw f(\bx, \bw) ^T \nabla_\bw f(\bx, \bw)) + \lambda I.
\end{equation}

\subsection{Long Horizon Exploration with Foundation Model}\label{sec:long-term}
We propose using a multimodal LLM as a zero-shot long-horizon planner and leaving the detailed path planning to a closed-form uncertainty-aware motion planning algorithm.

As we consider an agent moving in a 2D action plane (e.g., a ground robot), we leverage the view synthesis ability of 3DGS to create expressive bird's-eye view renderings that could provide an overview of the environment. 
An occupancy grid on the motion plane of our robot is created using our 3D Gaussian representation, which can be used to identify the frontiers of the current environment.
The frontiers are defined as points a set of neighboring points on the action plane on the boundary between free space and unobserved space. 
The agent has many choices of frontiers, especially during the early stages of exploration. 
We apply Chain-of-Thought Prompting~\cite{wei2022chain} to encourage the multimodal LLM to provide analysis on the candidate frontiers first before selection. 
We provide contextual information such as the description of the task, the total steps allowed for exploration, and our current step in the textual prompt.
The LLM is also allowed to decide whether the robot should not go to a frontier and instead focus on improving existing regions. %
We annotate the rendered map with the current location, previously visited trajectory and the frontiers which form the candidate long-term navigation goals. We prompt the LLM with both the contextual text information and annotated map as shown in \cref{fig:chat} and extract the long-term goal from its output.
By using a set of possible candidates we avoid possible infeasible destinations for motion planning. For example, the unexplored regions we have no reconstruction for at this point might be unreachable from the current space. 
We can also check that the LLM's output is actually one of the provided options before proceeding, if it is not then we instead select the largest frontier.
We use GPT-4o~\cite{gpt-4o} for our long-horizon planning task, but our method is agnostic to the underlying multimodal LLM as the inputs for the LLM are text prompts and a bird's-eye view rendering. 
Detailed examples can be found in the Supplementary.

\subsection{Expected Information Gain for Path Proposal and Selection}
\label{sec: path eig}
We use the Expected Information Gain (EIG) as both a preliminary method to further refine our candidate poses and as part of our path selection criteria. 

After identifying a long-horizon goal with the multimodal LLM, we form an initial set of candidate poses $\mathcal{T}_I$ by sampling points in the coarse region of interest from the LLM.
If there are no unvisited boundaries or the LLM suggests we do not need to explore the frontiers, we sample poses across the free space to form $\mathcal{T}_I$. 
We then evaluate the Expected Information Gain (EIG) for each pose $\bxa \in \mathcal{T}_I$, given by
\begin{equation}\label{eq:eig Gaussian}
    \mathtt{EIG}(\bxa) = \tr\left( \bH''[\bya | \bxa, \bw] \; \mathcal{I}(\bw)^{-1}\right),
\end{equation} 
as a preliminary selection metric to form our final candidate target poses set $\mathcal{T}_F$.
We calculate $\mathcal{I}(\bw)$ differently to FisherRF~\cite{jiang2024fisherrf}, which approximates it using $\bH''[\bw| D^{train}]$ by computing the Hessians on the training set.
This is also known as empirical Fisher Information, whose limitations have been widely discussed~\cite{kunstner2019limitations,martens2020new}. In most scenarios, this is a reluctant design choice because the distribution of $\bx \sim p(\bx)$ is unknown (i.e., the distribution of all possible images). However, $\bx \sim p(\bx)$ in our case is tractable because it represents the possible locations where we can take an observation for the environment, i.e. the free space of our map. 
Therefore, unlike FisherRF~\cite{jiang2024fisherrf} as described in Eq.~\ref{eq:agslam-opt}, we propose to use Monte-Carlo sampling to compute the Fisher Information of the current model
\begin{equation}
    \mathcal{I}(\bw) = \mathbb{E}_{\bx \sim p(\bx)} \left[ \bH''[\by| \bx, \bw] \right]  \simeq \sum_{k=1}^{N} \bH''[\by_k| \bx_k, \bw],%
\end{equation}
where $\bx_k$ is drawn from a uniform distribution of camera poses in the free space of the current map which we use to approximate $p$. We also uniformly initialize 3D Gaussians in the space, which will be subsequently updated with rendering losses for visited regions.

After the set of final candidate poses $\mathcal{T}_F$ has been formed, we construct paths to each candidate pose with the A* algorithm~\cite{hart1968formal} using the occupancy map. 
The path can be defined as an ordered set of camera poses from the current location $x_t$ at exploration step $t$ to the frontier points $x_{T}^j$ from the long horizon planning.

\begin{equation}
    \mathcal{P}_j = \{x_{t+1}^j, \dots,  x_{T}^j\}
\end{equation}

To evaluate how beneficial a path $\mathcal{P}_j$ would be for improving the reconstruction of the scene we consider the EIG for the 3D Gaussian parameters along that path, which can be computed as the sum over the path of the following term~\cite{jiang2024fisherrf}:
\begin{equation}
   \mathtt{EIG}_{\mathcal{P}_j, i}(\bx_i) = \tr\left( \bH''[\by_i | \bx_i, \bw] \; \mathcal{I}_{\mathcal{P}_j, i}(\bw)^{-1}
   \right)
\end{equation}
where $\mathcal{I}_{\mathcal{P}_j, i}(\bw)$ takes the mutual information along the path into account as follows
\begin{equation}
    \mathcal{I}_{\mathcal{P}_j, i}(\bw) = \mathcal{I}(\bw) + \sum_{\bx_t \in \mathcal{P}_j, t < i} \mathcal{I}(\bw|\bx_t).
\end{equation}

\begin{figure*}[t]
\setlength{\tabcolsep}{0pt}
	\centering
\begin{tabular}{cccccc}
ANS~\cite{chaplot2020learning} & Active-INR~\cite{yan2023active-neural-mapping} & UPEN~\cite{upen} & ExplORB~\cite{placed2022explorb} & FBE~\cite{yamauchi1997frontier} &  Ours \\ 
\includegraphics[width=0.16\textwidth,trim={2.5cm 2.5cm 2.5cm 2.5cm},clip]{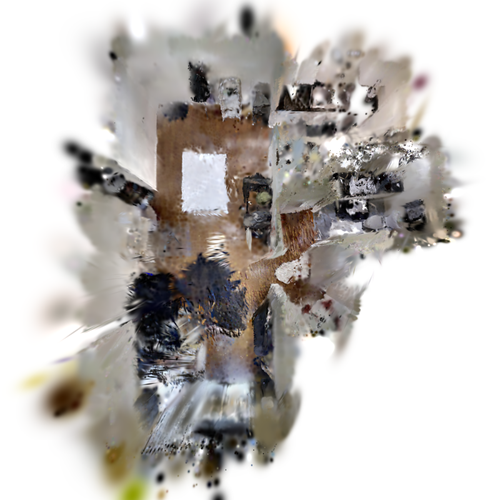} &
\includegraphics[width=0.16\textwidth,trim={2.5cm 2.5cm 2.5cm 2.5cm},clip]{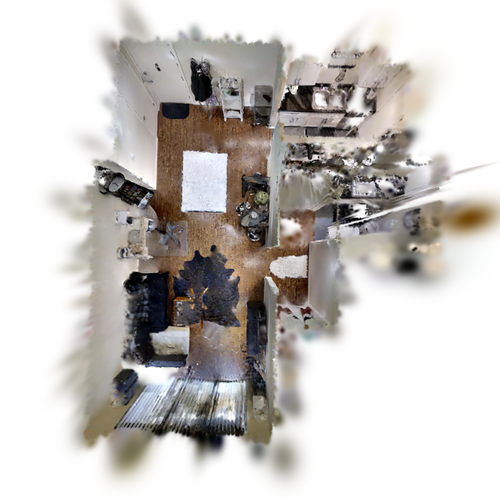} &
\includegraphics[width=0.16\textwidth,trim={2.5cm 2.5cm 2.5cm 2.5cm},clip]{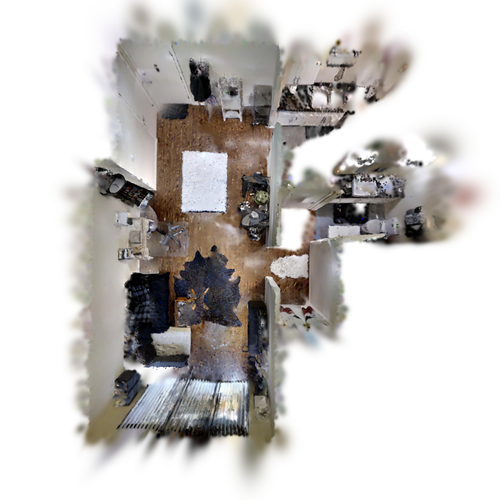} &
\includegraphics[width=0.16\textwidth,trim={2.5cm 2.5cm 2.5cm 2.5cm},clip]{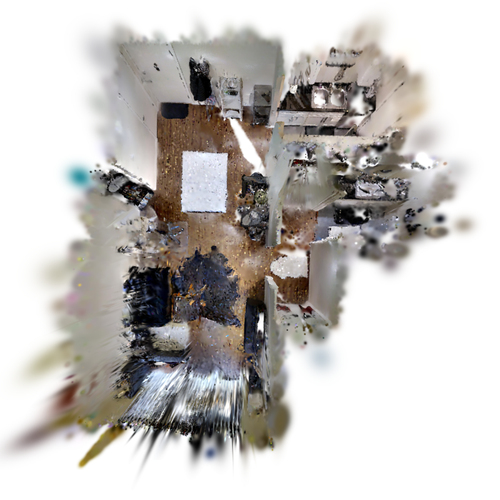} &
\includegraphics[width=0.16\textwidth,trim={2.5cm 2.5cm 2.5cm 2.5cm},clip]{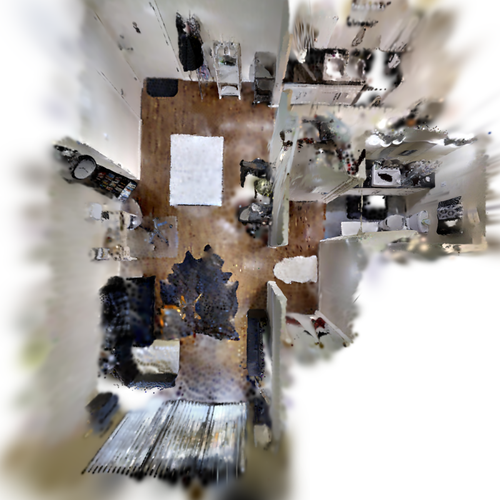} &
\includegraphics[width=0.16\textwidth,trim={2.5cm 2.5cm 2.5cm 2.5cm},clip]{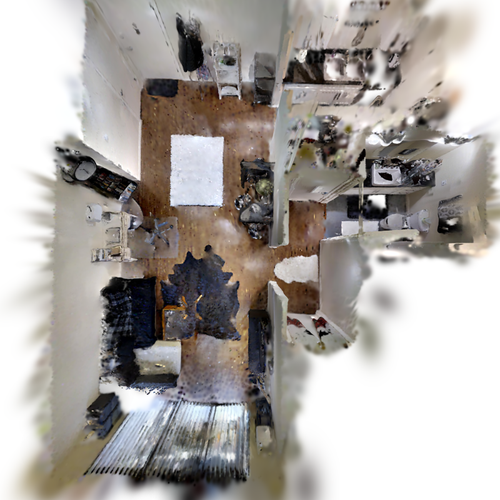} 
\\
\includegraphics[width=0.16\textwidth,trim={0cm 5cm 0cm 0cm},clip]{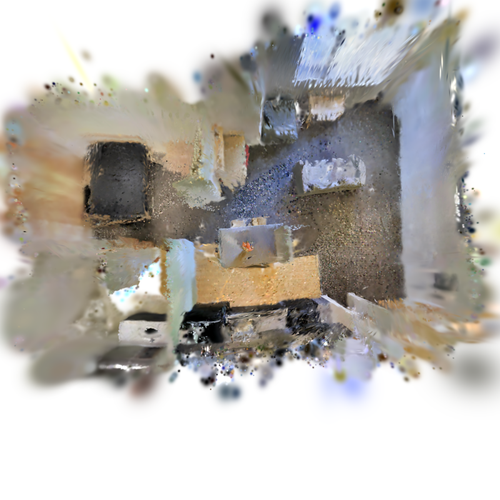} &
\includegraphics[width=0.16\textwidth,trim={0cm 5cm 0cm 0cm},clip]{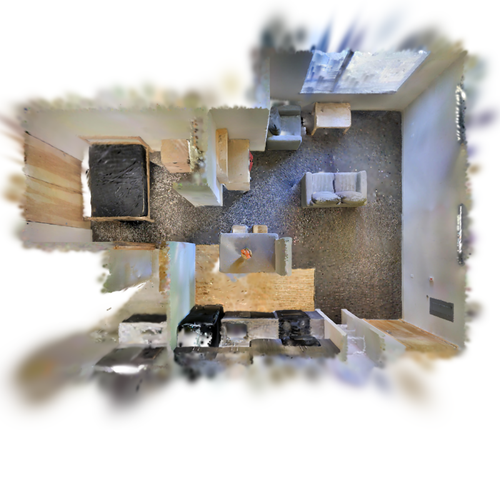} &
\includegraphics[width=0.16\textwidth,trim={0cm 5cm 0cm 0cm},clip]{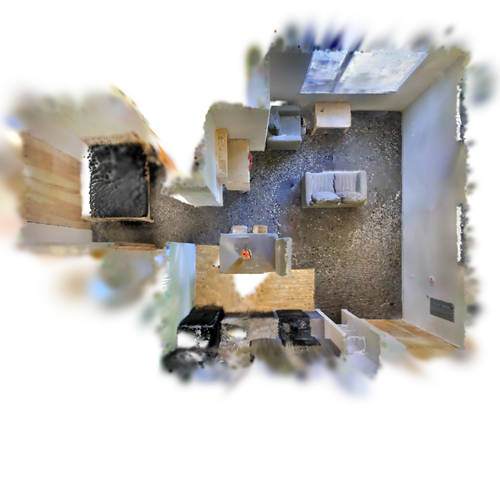} &
\includegraphics[width=0.16\textwidth,trim={0cm 5cm 0cm 0cm},clip]{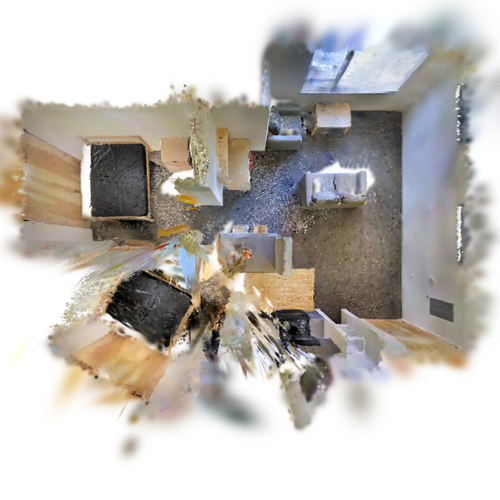} &
\includegraphics[width=0.16\textwidth,trim={0cm 5cm 0cm 0cm},clip]{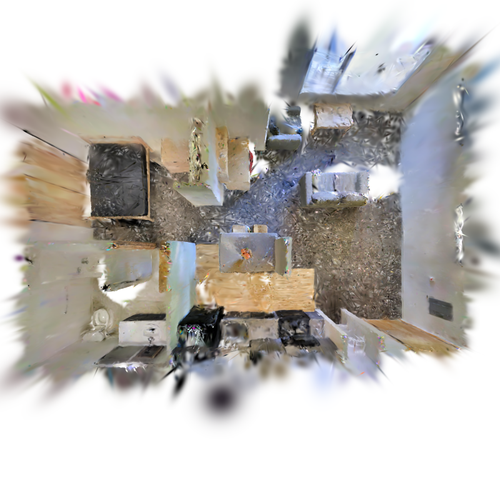} &
\includegraphics[width=0.16\textwidth,trim={0cm 5cm 0cm 0cm},clip]{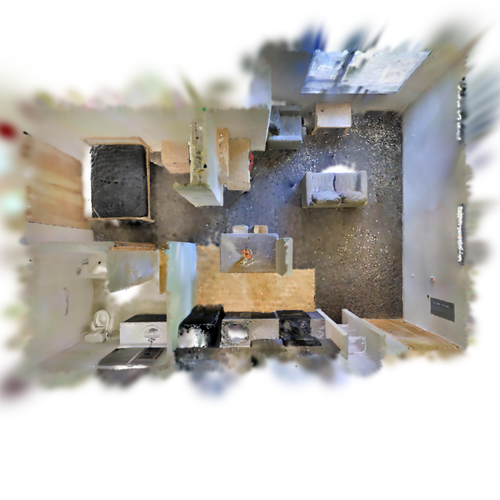} 
\end{tabular}
\caption{{\bf Qualitative Comparison for Final Scene Reconstruction on Gibson Dataset} Greigsville (top) and Ribera (bottom) scenes. We provide top-down rendering for different methods. Note that UPEN and Active-INR use GT pose in this visualization.}\label{fig:qual-gibson}
\end{figure*}

\begin{figure*}[t]
\setlength{\tabcolsep}{0pt}
	\centering
\begin{tabular}{cccccc}
UPEN~\cite{upen} & FBE~\cite{yamauchi1997frontier} & Ours & UPEN~\cite{upen} & FBE~\cite{yamauchi1997frontier} & Ours \\ 
\includegraphics[width=0.16\textwidth]{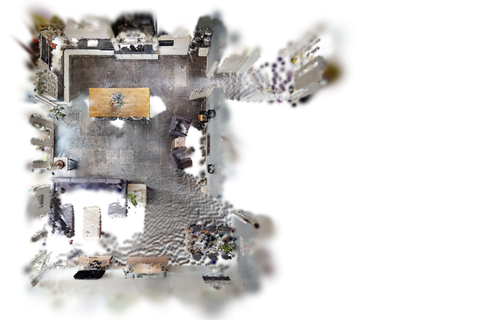} &
\includegraphics[width=0.16\textwidth]{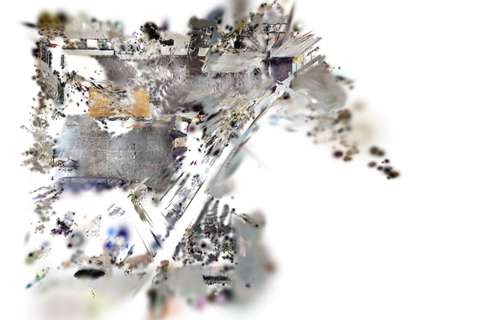} &
\includegraphics[width=0.16\textwidth]{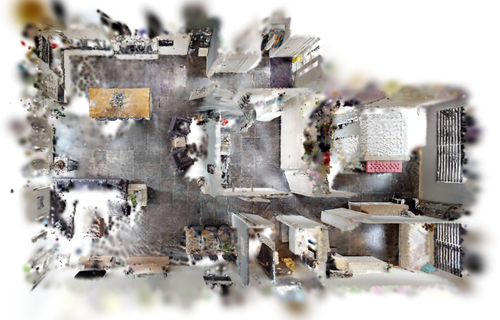} &
\includegraphics[width=0.16\textwidth]{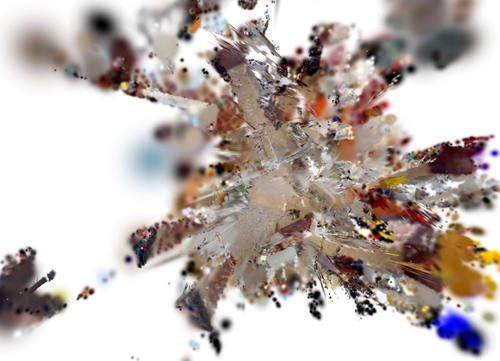} &
\includegraphics[width=0.16\textwidth]{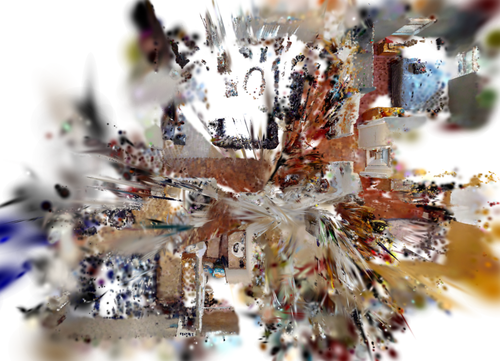} &
\includegraphics[width=0.16\textwidth]{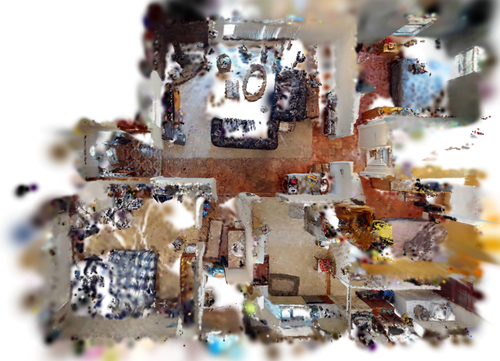} 
\end{tabular}
\caption{{\bf Qualitative Comparison for Final Scene Reconstruction on Habitat-Matterport 3D Dataset} oPj9qMxrDEa~(left) and QKGMrurUVbk~(right) scenes. We provide top-down rendering for different methods. }\label{fig:qual-hm3d}
\end{figure*}

\subsection{Localization Uncertainty for Path Selection}\label{sec:path-sample}
\label{ssec:path localization}

If solely maximizing the EIG, the robot will be more likely to explore unvisited regions.
However, exploring regions that have not been well reconstructed also means the agent would have the risk of worse localization accuracy due to noise and ambiguities in the unreconstructed regions during pose optimization.
The cost of localization must thus be considered during path planning to balance the importance of exploring new environments with maintaining localization accuracy. 
We propose to use Fisher Information as a measurement for localization uncertainty to address this. During optimization, we essentially optimize on the logarithmic mapping $\tau_i \triangleq \Log(\bx_i)$ of our camera pose.
By the Cramér–Rao bound, the covariance of $\tau_i \in \se{3}$ can be lower-bounded with the inverse of Fisher Information matrix $\mathcal{I}(\tau_i)$:
\begin{equation}
    \Cov(T(\hat{\tau_i})) \geq \mathcal{I}(\tau_i)^{-1}
\end{equation}
where $T(\tau_i)$ is an unbiased estimator for $\tau_i$ solved by iteratively optimizing photo-metric loss. 
Hence, we can define the localization cost $C_{\textit{local}}$ at a pose $\bx_i$ in terms of $\tau_i$ as:
\begin{equation}
    C_{\textit{local}}(\tau_i) = -\log\det (\nabla_{\tau_i} f(\tau_i, \bw)^T \nabla_{\tau_i} f(\tau_i, \bw))
\end{equation}

Matsuki~\etal~\cite{matsuki2023gaussian} computed the Jacobians of the camera pose with respect to the mean and covariances of each gaussian $\pd{\meanI}{\camCW}$ and $\pd{\covI}{\camCW}$. However, we need to compute the Jacobian of $\tau_i$
with respect to the rendering output:
\begin{equation}
    \nabla_{\tau_i} f(\tau_i, \bw) = \pd{f(\tau_i, \bw)}{\tau_i} = \begin{bmatrix}
        \pd{f(\tau_i, \bw)}{\meanI} & \pd{f(\tau_i, \bw)}{\covI}
    \end{bmatrix} \begin{bmatrix}
         \mpd{\meanC}{\tau_i} \\  \mpd{\matW}{\tau_i}
    \end{bmatrix}
\end{equation}
Unlike 3DGS parameters, our computation of the localization cost can be directly computed on the full Fisher Information matrix without using the Laplace approximation~\cite{laplace2021, bayesian-interpolation}. That is because the full Jacobian $\nabla_{\tau_i} f(\tau_i, \bw)$ is linear to the number of Gaussians.%

We select the best path by minimizing the total cost for all viewpoints $\bx_i$ along a path $\mathcal{P}_j$:
\begin{equation}\label{eq:agslam-path}
    \argmin_{\mathcal{P}_j} \sum_{\bx_i \in \mathcal{P}_j} C_{\textit{local}}(\Log(\bx_i)) - \eta \log(\mathtt{EIG}_{\mathcal{P}_j,i}(\bx_i))
\end{equation}
where $\eta$ is a hyper-parameter controlling the relative importance between EIG and localization accuracy. %
Our active mapping system constantly updates the map, and we replan using our active path planning algorithm if we detect the agent is getting close to a possible obstacle or upon reaching the end of the previously selected path.

\section{Experiments}

\begin{table*}[t]
  \centering
  
      \begin{tabular}{l|llllll}
        \toprule
        Method & PSNR $\uparrow$ & SSIM $\uparrow$ & LPIPS $\downarrow$ & Depth MAE $\downarrow$ & RMSE ATE $\downarrow$ & Completeness (\%) $\uparrow$
        \\
        \midrule
        ANS~\cite{chaplot2020learning} & 16.34 & 0.6818 & 0.3923 & 0.3886 & 0.1105 &  35.33
        \\ \midrule
        UPEN~\cite{upen}  & 16.44 & 0.6678  & 0.4134 & 0.4841 & 0.5158 & 22.66 
        \\
        \midrule
        ExplORB~\cite{placed2022explorb} & 18.99 & 0.7175 & 0.3994 & 0.2664 & 0.2296 & 30.23 
        \\
        \midrule
        FBE~\cite{yamauchi1997frontier} & 21.45 & 0.7618 &  \textbf{0.2126} &  0.1028 & 0.1680 & 55.87
        \\
        \midrule
        Ours & \textbf{23.28} & \textbf{0.8067} & 0.2507 & \textbf{0.0696} & \textbf{0.0226} & \textbf{84.38} 
        \\
        \bottomrule
      \end{tabular}
  \caption{\textbf{Quantitative Evaluation on Reconstruction Quality and Tracking Accuracy on Gibson dataset}}\label{tab:results gibson}
\end{table*}

\begin{table*}[t]
  \centering
  
      \begin{tabular}{l|llllllll}
        \toprule
        Method & PSNR $\uparrow$ & SSIM $\uparrow$ & LPIPS $\downarrow$ & Depth MAE $\downarrow$ & RMSE ATE $\downarrow$  & Completeness (\%) $\uparrow$
        \\
        \midrule
        UPEN~\cite{upen}  & 12.23 & 0.4795  & 0.5157 & 0.7356 & 0.4393 & 17.48 
        \\
        \midrule
        ExplORB~(gt)~\cite{placed2022explorb}  &  17.81 & 0.3694 & 0.6810 & 0.5071 & - & 31.92
        \\
        \midrule
        FBE~\cite{yamauchi1997frontier} & 15.80 & 0.5952 &  0.4392 & 0.4085 & 1.2004 & 22.42 
        \\
        \midrule
        Ours & \textbf{19.86}  & \textbf{0.7127} & \textbf{0.4122} & \textbf{0.1666} & \textbf{0.0336} & \textbf{49.76}
        \\
        \bottomrule
      \end{tabular}
  \caption{\textbf{Quantitative Evaluation on Reconstruction Quality and Tracking Accuracy on HM3D dataset}}\label{tab:results hm3d}
\end{table*}

\begin{table*}[t]
  \centering
  \resizebox{.98\textwidth}{!}{
  
      \begin{tabular}{l|lllllll}
        \toprule
        Method & PSNR $\uparrow$ & SSIM $\uparrow$ & LPIPS $\downarrow$ & Depth MAE $\downarrow$ & RMSE ATE $\downarrow$  & Completeness (\%) $\uparrow$
        \\
        \midrule
        w.o. LLM \& Localization Uncertainty & 16.15 & 0.6550 & 0.6193 & 0.3409  & 0.2478 & 35.40
        \\ \midrule
        w.o. LLM & 16.94 & 0.6799 & 0.5847 & 0.2887	& 0.1694 & 37.26
        \\ \midrule
        Ours~(Llava-7b) & 18.46 & 0.6805 & 0.4623 &  0.2033 & \textbf{0.0159} & 17.41
        \\ \midrule
        $\lambda=2\times 10^{-6}$~$\dagger$ & 18.90 &	0.6976	&	0.4408 & 0.1966		 & 0.0479  & 18.12
        \\ \midrule
        $\lambda=5\times 10^{-6}$~$\dagger$ &  18.90 &	0.6946 &	0.4602 &	0.2145 & 0.0551	& 18.05	
        \\ \midrule
         Ours$\dagger$ & \textbf{19.86}  & \textbf{0.7127} & \textbf{0.4122} & \textbf{0.1666} & \textbf{0.0336} & \textbf{49.76}
        \\
        \bottomrule
      \end{tabular}
  }
  \caption{\textbf{Ablation Study of Localization Uncertainty Term on Scenes from the HM3D Dataset.} We compare our method with and without the localization uncertainty term to validate that including it improves localization and reconstruction.
  $\dagger$: the multimodal LLM is GPT-4o in the experiment.}
  \label{tab:abl-pose-uncertainty}
\end{table*}

\subsection{Experimental Set-up}
\paragraph{Dataset}
\label{ssec: dataset}
Following previous methods~\cite{upen, yan2023active-neural-mapping}, our algorithm is evaluated in the Habitat Simulator~\cite{szot2021habitat} on the Gibson~\cite{xiazamirhe2018gibsonenv} and Habitat-Matterport 3D (HM3D)~\cite{ramakrishnan2021hm3d}
datasets comprised of indoor scenes reconstructed from scans of real houses. %
We adopt the default start point in the Habitat Simulator as the starting point for exploration in each scene. The total number of steps for each experiment is 2000. The system takes color and depth images at the resolution of 800x800 and outputs a discrete action at each step. The action space consists of MOVE FORWARD by 5cm, TURN LEFT, and TURN RIGHT by 5$\degree$. The field of view (FOV) is set to 90$\degree$ vertically and horizontally. 
Please refer to the supplement for more details about the evaluation split and other hyper-parameters.\label{ssec: compute}

\paragraph{Metrics}
We evaluated our method using the Peak-signal-to-noise ratio~(PSNR), Structural Similarity Index Measure ~(SSIM)~\cite{wang2004image}, Learned Perceptual Image Patch Similarity (LPIPS)~\cite{zhang2018unreasonable} for RGB rendering and mean absolute error~(MAE) for depth rendering as metrics for scene reconstruction quality. 
We calculate these metrics using 2000 points uniformly sampled from the movement plane of the agent in the scene, discarding any points that are not navigable.
We argue that the rendering quality reflects both reconstruction quality and pose accuracy because high tracking accuracy would help the training of the 3D Gaussian Splatting model. Meanwhile, misaligned poses will lead to misaligned rendering at test time thus leading to inferior results. 
Following previous approaches~\cite{yan2023active-neural-mapping,kuang2024active}, we also use the completion ratio as an evaluation metric. 
To evaluate the pose estimation accuracy, we use the root mean squared average tracking error~(RMSE ATE), but as the trajectories for each method are different, the RMSE ATE should only be considered along with other metrics such as completness.

\paragraph{Baselines}
We compare to two exploration methods which assume ground truth pose: UPEN~\cite{upen} and Active Neural Mapping (active-INR)~\cite{yan2023active-neural-mapping}. %
We also compare our method with Active Neural SLAM (ANS)~\cite{chaplot2020learning}, explORB~\cite{placed2022explorb} and Frontier Based Exploration (FBE)~\cite{yamauchi1997frontier} without ground truth pose provided. 
Note that we do not run ANS on HM3D as it is not trained on this dataset.

To compare the rendering quality fairly, we run all the baselines using the MonoGS~\cite{matsuki2023gaussian} backend for reconstruction. We run UPEN and FBE online, but for ANS, active-INR, and ExplORB, we record and playback trajectories obtained using their source code. 
Because the forward step size for ANS is much larger than for our method, we interpolate the trajectory so that the forward step size matches our method's to make the steps comparable.
For ExplORB, since the official implementation is based on MoveBase, which uses velocity commands, we sample the trajectory at 5 Hz. %
We also found that ANS, active-INR, and UPEN failed on some scenes due to localization failure of the MonoGS backend. ANS produces a pose estimate (using information from noisy pose sensors not provided to our pipeline), so we set the pose estimate of the MonoGS backend to the one from ANS. As active-INR and UPEN do not produce a pose estimate, we evaluate them using the ground-truth pose.

\begin{figure*}[!tbp]
  \begin{subfigure}[b]{0.6\textwidth}
    \includegraphics[width=\textwidth]{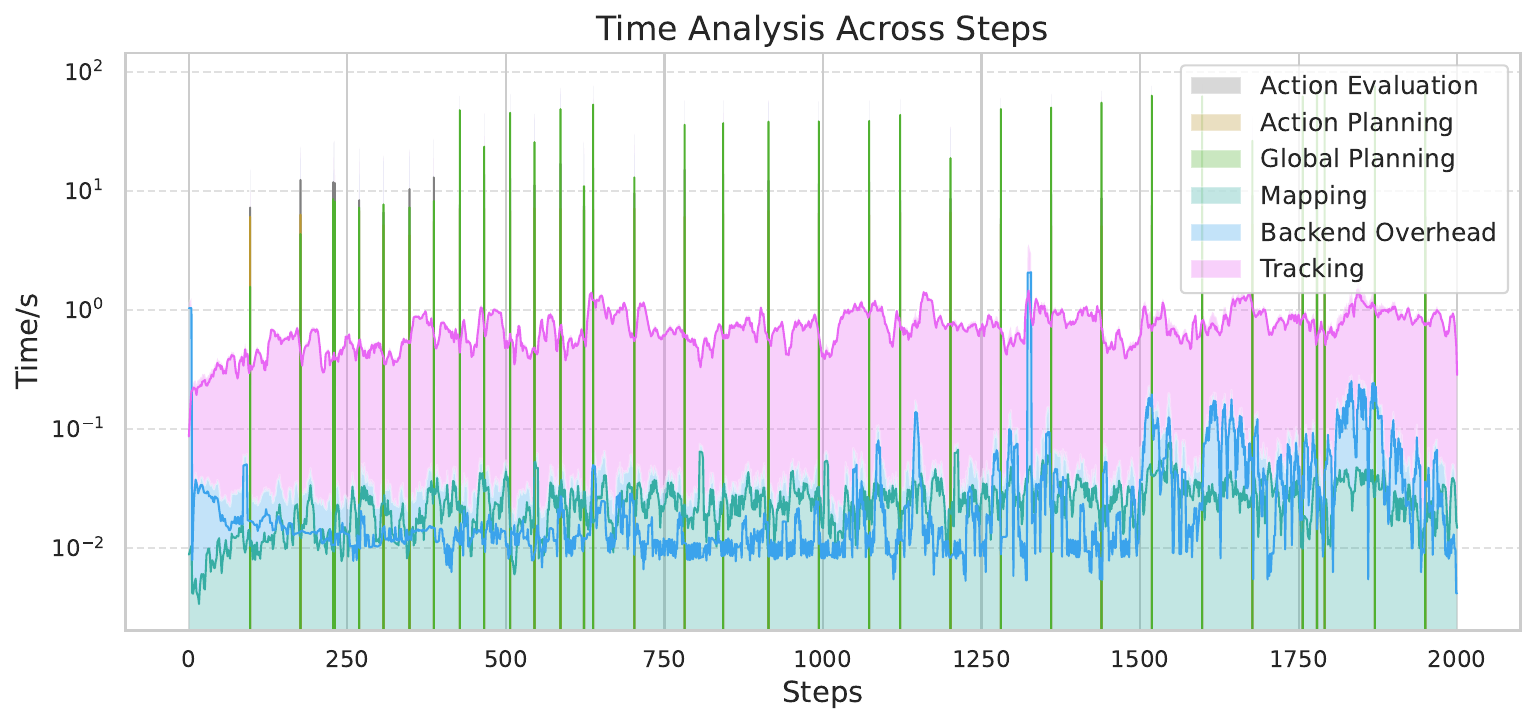}
  \end{subfigure}
  \hfill
  \begin{subfigure}[b]{0.4\textwidth}
    \includegraphics[width=\textwidth]{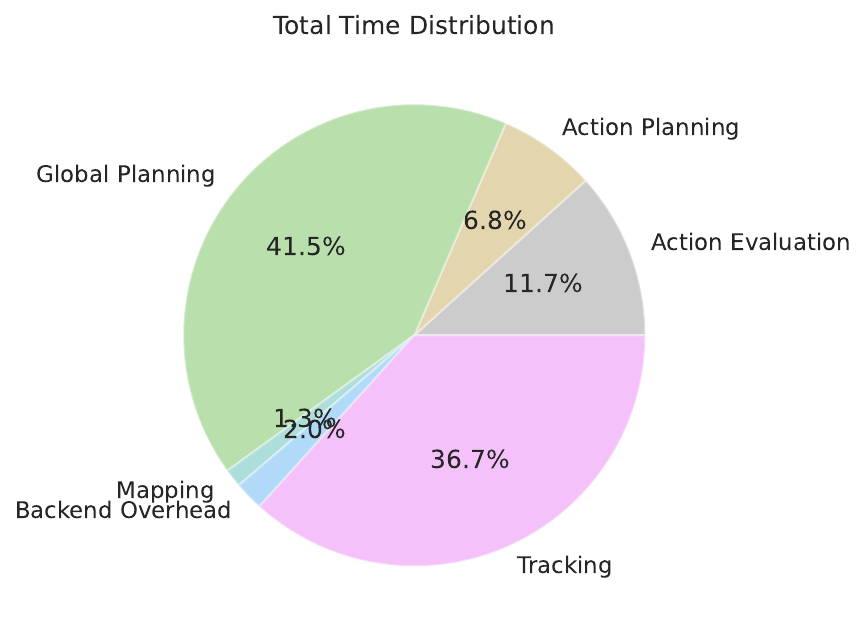}
  \end{subfigure}
  \caption{\textbf{Runtime Analysis of the Active Mapping System.} We provide running time statistics by each step and by the percentage of total time consumed on one episode.}\label{fig:time-analysis}
\end{figure*}

\begin{figure}[t]
\setlength{\tabcolsep}{0pt}
	\centering
\includegraphics[width=0.5\textwidth, trim={1.8cm 6cm 2.2cm 2.cm},clip]{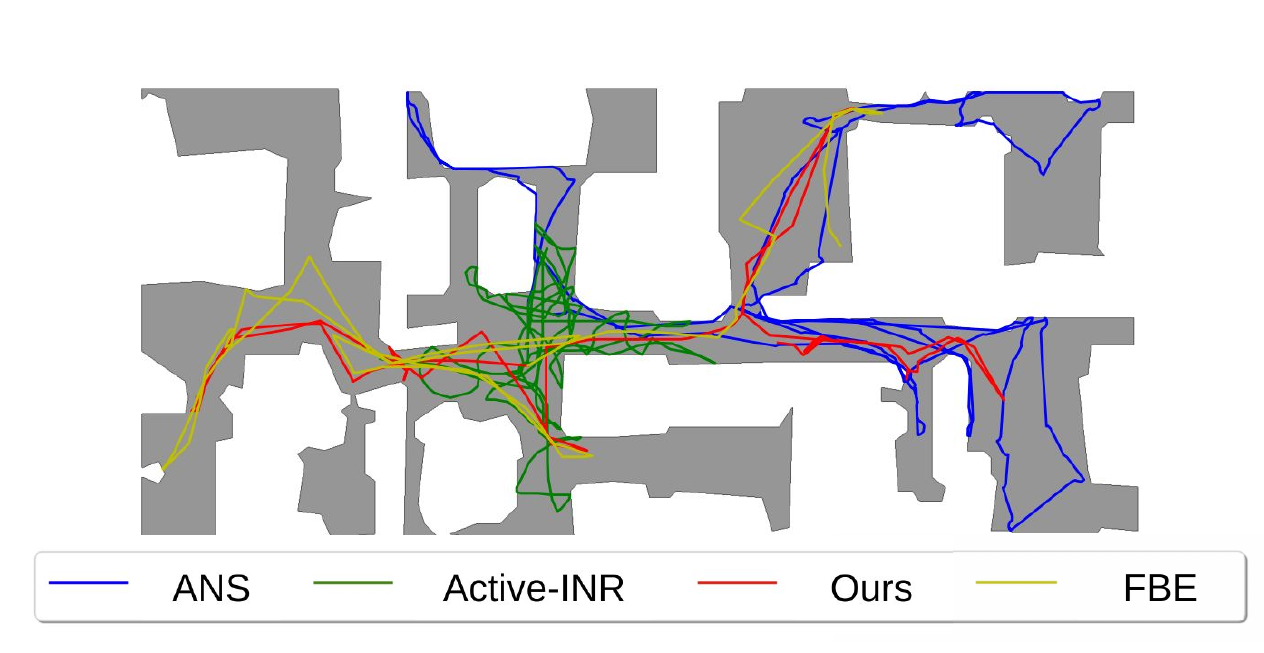} 
\caption{{\bf Qualitative Comparison for Trajectories} on Cantwell from Gibson Dataset.}\label{fig:qual-traj}
\end{figure}

\subsection{Comparison Against Previous Methods}
\label{ssec: results}
Table~\ref{tab:results gibson} shows the results of our method and the baselines for exploration in scenes from the Gibson dataset~\cite{xiazamirhe2018gibsonenv}, and Table~\ref{tab:results hm3d} shows the results on HM3D~\cite{Matterport3D}. %
Our method outperforms the baselines on all metrics. %
We further qualitatively compare the reconstruction qualities after active exploration in Fig.~\ref{fig:qual-gibson} and Fig.~\ref{fig:qual-hm3d}, and the trajectories in Fig.~\ref{fig:qual-traj}. Our method does not have major errors from failed localization, and we have fewer gaps in the scenes than other methods. For example, in the Ribera scene, all methods except for us and FBE miss the bathroom at the bottom left, and FBE misses more of the area around the sofa than us. %
We show the trajectories taken for the Cantwell scene from the Gibson dataset, which is a relatively large and challenging scene. We show only a few baselines to keep the figure legible. %
We can see that ANS~\cite{chaplot2020learning} does not go to the region on the bottom left, Active-INR~\cite{yan2023active-neural-mapping} stays in the center region, and FBE does not go to the region on the bottom right, whereas our method efficiently visits most areas of the scene.

\subsection{Ablation Study and Runtime Analysis}

To validate the effectiveness of localization uncertainty, hyperparameters, and LLM, we performed an ablative study of each component in our system on the HM3D dataset in Table.~\ref{tab:abl-pose-uncertainty}. As can be seen, the average trajectory error is much lower with the localization uncertainty than without. In addition, after adding LLM, the model performs better in all metrics. We also noticed the tracking error was lower after introducing multimodal LLM to the active mapping system. We hypothesize this improvement is because improved scene reconstruction can also help reduce tracking errors.

We additionally provide a breakdown of the runtime of our method, shown in~\cref{fig:time-analysis}.

\section{Conclusion}
\label{sec: conclusion}
We present an active mapping system that can autonomously explore an environment, the first method to do so using a 3D Gaussian representation without ground truth camera poses. By rendering maps using our scene representation, we can seamlessly query a multimodal Large Language Model for long-horizon planning to leverage its knowledge of scene layouts. This allows the robot to efficiently explore the scene while avoiding setting unreachable navigation goals. We then plan an optimal path to the long-horizon goal using 
our information-driven path proposal and selection algorithm, which balances the information gain with respect to the acene against the localization uncertainty, ensuring that the robot captures the geometric details of the scene for a high-quality reconstruction. We evaluate our method on scenes from the Gibson~\cite{xiazamirhe2018gibsonenv} and Habitat-Matterport 3D~\cite {ramakrishnan2021hm3d} datasets in terms of the rendering quality, completeness, and average tracking error.
To enable our method to support more robotics applications, future work could extend it to take advantage of our 3D scene representation and consider movement with higher degrees of freedom (DOF) than the currently supported 3DOF. Our method could also be extended to task-driven exploration such as finding objects~\cite{batra2020objectnav,khanna2024goat}, leveraging the LLM for efficient search. Incorporating semantic features~\cite{zhou2024feature, shi2024language, qin2024langsplat} to allow for grounding language to the scene would also enable many robotics and computer vision applications.

\clearpage
{\small
\paragraph{Acknowledgements}
The authors gratefully appreciate support through the following grants: NSF FRR 2220868, NSF IIS-RI 2212433, NSF TRIPODS 1934960, NSF CPS 2038873.
The authors appreciate the help from Prof. Pratik Chaudhari for the insightful discussion and hardware platforms.
}
{
    \small
    \bibliographystyle{ieeenat_fullname}
    \bibliography{main}

\begin{thebibliography}{98}
\providecommand{\natexlab}[1]{#1}
\providecommand{\url}[1]{\texttt{#1}}
\expandafter\ifx\csname urlstyle\endcsname\relax
  \providecommand{\doi}[1]{doi: #1}\else
  \providecommand{\doi}{doi: \begingroup \urlstyle{rm}\Url}\fi

\bibitem[Asgharivaskasi and Atanasov(2023)]{Semantic-OcTree}
Arash Asgharivaskasi and Nikolay Atanasov.
\newblock Semantic octree mapping and shannon mutual information computation for robot exploration.
\newblock \emph{IEEE Transactions on Robotics}, 39\penalty0 (3):\penalty0 1910--1928, 2023.

\bibitem[Batra et~al.(2020)Batra, Gokaslan, Kembhavi, Maksymets, Mottaghi, Savva, Toshev, and Wijmans]{batra2020objectnav}
Dhruv Batra, Aaron Gokaslan, Aniruddha Kembhavi, Oleksandr Maksymets, Roozbeh Mottaghi, Manolis Savva, Alexander Toshev, and Erik Wijmans.
\newblock {ObjectNav} {R}evisited: {O}n {E}valuation of {E}mbodied {A}gents {N}avigating to {O}bjects.
\newblock \emph{arXiv preprint arXiv:2006.13171}, 2020.

\bibitem[Brohan et~al.(2023)Brohan, Chebotar, Finn, Hausman, Herzog, Ho, Ibarz, Irpan, Jang, Julian, et~al.]{brohan2023can}
Anthony Brohan, Yevgen Chebotar, Chelsea Finn, Karol Hausman, Alexander Herzog, Daniel Ho, Julian Ibarz, Alex Irpan, Eric Jang, Ryan Julian, et~al.
\newblock Do as i can, not as i say: Grounding language in robotic affordances.
\newblock In \emph{Conference on robot learning}, pages 287--318. PMLR, 2023.

\bibitem[Bry and Roy(2011)]{bry2011rapidly}
Adam Bry and Nicholas Roy.
\newblock Rapidly-exploring random belief trees for motion planning under uncertainty.
\newblock In \emph{2011 IEEE international conference on robotics and automation}, pages 723--730. IEEE, 2011.

\bibitem[Campos et~al.(2021)Campos, Elvira, Rodr{\'\i}guez, Montiel, and Tard{\'o}s]{campos2021orb}
Carlos Campos, Richard Elvira, Juan J~G{\'o}mez Rodr{\'\i}guez, Jos{\'e}~MM Montiel, and Juan~D Tard{\'o}s.
\newblock Orb-slam3: An accurate open-source library for visual, visual--inertial, and multimap slam.
\newblock \emph{IEEE Transactions on Robotics}, 37\penalty0 (6):\penalty0 1874--1890, 2021.

\bibitem[Carlone et~al.(2014)Carlone, Du, Kaouk~Ng, Bona, and Indri]{carlone2014active}
Luca Carlone, Jingjing Du, Miguel Kaouk~Ng, Basilio Bona, and Marina Indri.
\newblock Active slam and exploration with particle filters using kullback-leibler divergence.
\newblock \emph{Journal of Intelligent \& Robotic Systems}, 75:\penalty0 291--311, 2014.

\bibitem[Carrillo et~al.(2012)Carrillo, Reid, and Castellanos]{carrillo2012comparison}
Henry Carrillo, Ian Reid, and Jos{\'e}~A Castellanos.
\newblock On the comparison of uncertainty criteria for active slam.
\newblock In \emph{2012 IEEE International Conference on Robotics and Automation}, pages 2080--2087. IEEE, 2012.

\bibitem[Chang et~al.(2017)Chang, Dai, Funkhouser, Halber, Niessner, Savva, Song, Zeng, and Zhang]{Matterport3D}
Angel Chang, Angela Dai, Thomas Funkhouser, Maciej Halber, Matthias Niessner, Manolis Savva, Shuran Song, Andy Zeng, and Yinda Zhang.
\newblock Matterport3d: Learning from rgb-d data in indoor environments.
\newblock \emph{International Conference on 3D Vision}, 2017.

\bibitem[Chaplot et~al.(2020)Chaplot, Gandhi, Gupta, Gupta, and Salakhutdinov]{chaplot2020learning}
Devendra~Singh Chaplot, Dhiraj Gandhi, Saurabh Gupta, Abhinav Gupta, and Ruslan Salakhutdinov.
\newblock Learning to explore using active neural slam.
\newblock In \emph{ICLR}, 2020.

\bibitem[Chen et~al.(2024{\natexlab{a}})Chen, Lin, Liu, Ma, Liang, and Wong]{chen2024affordances}
Jiaqi Chen, Bingqian Lin, Xinmin Liu, Lin Ma, Xiaodan Liang, and Kwan-Yee~K Wong.
\newblock Affordances-oriented planning using foundation models for continuous vision-language navigation.
\newblock \emph{arXiv preprint arXiv:2407.05890}, 2024{\natexlab{a}}.

\bibitem[Chen et~al.(2024{\natexlab{b}})Chen, Lin, Xu, Chai, Liang, and Wong]{chen2024mapgpt}
Jiaqi Chen, Bingqian Lin, Ran Xu, Zhenhua Chai, Xiaodan Liang, and Kwan-Yee Wong.
\newblock Mapgpt: Map-guided prompting with adaptive path planning for vision-and-language navigation.
\newblock In \emph{Proceedings of the 62nd Annual Meeting of the Association for Computational Linguistics (Volume 1: Long Papers)}, pages 9796--9810, 2024{\natexlab{b}}.

\bibitem[Chen et~al.(2019)Chen, Gupta, and Gupta]{chen2018learning}
Tao Chen, Saurabh Gupta, and Abhinav Gupta.
\newblock Learning exploration policies for navigation.
\newblock In \emph{International Conference on Learning Representations}, 2019.

\bibitem[Chen et~al.(2020)Chen, Huang, and Fitch]{chen2020active}
Yongbo Chen, Shoudong Huang, and Robert Fitch.
\newblock Active slam for mobile robots with area coverage and obstacle avoidance.
\newblock \emph{IEEE/ASME Transactions on Mechatronics}, 25\penalty0 (3):\penalty0 1182--1192, 2020.

\bibitem[Daxberger et~al.(2021)Daxberger, Kristiadi, Immer, Eschenhagen, Bauer, and Hennig]{laplace2021}
Erik Daxberger, Agustinus Kristiadi, Alexander Immer, Runa Eschenhagen, Matthias Bauer, and Philipp Hennig.
\newblock Laplace redux--effortless {B}ayesian deep learning.
\newblock In \emph{{N}eur{IPS}}, 2021.

\bibitem[Dorbala et~al.(2023)Dorbala, Mullen~Jr, and Manocha]{dorbala2023can}
Vishnu~Sashank Dorbala, James~F Mullen~Jr, and Dinesh Manocha.
\newblock Can an embodied agent find your “cat-shaped mug”? llm-based zero-shot object navigation.
\newblock \emph{IEEE Robotics and Automation Letters}, 2023.

\bibitem[Driess et~al.(2023)Driess, Xia, Sajjadi, Lynch, Chowdhery, Ichter, Wahid, Tompson, Vuong, Yu, Huang, Chebotar, Sermanet, Duckworth, Levine, Vanhoucke, Hausman, Toussaint, Greff, Zeng, Mordatch, and Florence]{driess2023palme}
Danny Driess, Fei Xia, Mehdi S.~M. Sajjadi, Corey Lynch, Aakanksha Chowdhery, Brian Ichter, Ayzaan Wahid, Jonathan Tompson, Quan Vuong, Tianhe Yu, Wenlong Huang, Yevgen Chebotar, Pierre Sermanet, Daniel Duckworth, Sergey Levine, Vincent Vanhoucke, Karol Hausman, Marc Toussaint, Klaus Greff, Andy Zeng, Igor Mordatch, and Pete Florence.
\newblock Palm-e: An embodied multimodal language model.
\newblock In \emph{arXiv preprint arXiv:2303.03378}, 2023.

\bibitem[Du et~al.(2023)Du, Yang, Florence, Xia, Wahid, Ichter, Sermanet, Yu, Abbeel, Tenenbaum, et~al.]{du2023video}
Yilun Du, Mengjiao Yang, Pete Florence, Fei Xia, Ayzaan Wahid, Brian Ichter, Pierre Sermanet, Tianhe Yu, Pieter Abbeel, Joshua~B Tenenbaum, et~al.
\newblock Video language planning.
\newblock \emph{arXiv preprint arXiv:2310.10625}, 2023.

\bibitem[Engel et~al.(2014)Engel, Sch{\"o}ps, and Cremers]{engel2014lsd}
Jakob Engel, Thomas Sch{\"o}ps, and Daniel Cremers.
\newblock Lsd-slam: Large-scale direct monocular slam.
\newblock In \emph{European conference on computer vision}, pages 834--849. Springer, 2014.

\bibitem[Engel et~al.(2017)Engel, Koltun, and Cremers]{engel2017direct}
Jakob Engel, Vladlen Koltun, and Daniel Cremers.
\newblock Direct sparse odometry.
\newblock \emph{IEEE transactions on pattern analysis and machine intelligence}, 40\penalty0 (3):\penalty0 611--625, 2017.

\bibitem[Ester et~al.(1996)Ester, Kriegel, Sander, Xu, et~al.]{ester1996density}
Martin Ester, Hans-Peter Kriegel, J{\"o}rg Sander, Xiaowei Xu, et~al.
\newblock A density-based algorithm for discovering clusters in large spatial databases with noise.
\newblock In \emph{KDD}, pages 226--231, 1996.

\bibitem[Feder et~al.(1999)Feder, Leonard, and Smith]{feder1999adaptive}
Hans Jacob~S Feder, John~J Leonard, and Christopher~M Smith.
\newblock Adaptive mobile robot navigation and mapping.
\newblock \emph{The International Journal of Robotics Research}, 18\penalty0 (7):\penalty0 650--668, 1999.

\bibitem[Feng et~al.(2024)Feng, Zhan, Chen, Yan, Xu, Cai, Li, Zhu, and Xu]{feng2024naruto}
Ziyue Feng, Huangying Zhan, Zheng Chen, Qingan Yan, Xiangyu Xu, Changjiang Cai, Bing Li, Qilun Zhu, and Yi Xu.
\newblock Naruto: Neural active reconstruction from uncertain target observations.
\newblock In \emph{Proceedings of the IEEE/CVF Conference on Computer Vision and Pattern Recognition}, pages 21572--21583, 2024.

\bibitem[Georgakis et~al.(2022)Georgakis, Bucher, Arapin, Schmeckpeper, Matni, and Daniilidis]{upen}
Georgios Georgakis, Bernadette Bucher, Anton Arapin, Karl Schmeckpeper, Nikolai Matni, and Kostas Daniilidis.
\newblock Uncertainty-driven planner for exploration and navigation.
\newblock In \emph{ICRA}, 2022.

\bibitem[Goli et~al.(2023)Goli, Reading, Sellán, Jacobson, and Tagliasacchi]{goli2023}
Lily Goli, Cody Reading, Silvia Sellán, Alec Jacobson, and Andrea Tagliasacchi.
\newblock {Bayes' Rays}: Uncertainty quantification in neural radiance fields.
\newblock \emph{arXiv}, 2023.

\bibitem[Hart et~al.(1968)Hart, Nilsson, and Raphael]{hart1968formal}
Peter~E Hart, Nils~J Nilsson, and Bertram Raphael.
\newblock A formal basis for the heuristic determination of minimum cost paths.
\newblock \emph{IEEE transactions on Systems Science and Cybernetics}, 4\penalty0 (2):\penalty0 100--107, 1968.

\bibitem[He et~al.(2010)He, Brunskill, and Roy]{he2010puma}
Ruijie He, Emma Brunskill, and Nicholas Roy.
\newblock Puma: Planning under uncertainty with macro-actions.
\newblock In \emph{Proceedings of the AAAI Conference on Artificial Intelligence}, pages 1089--1095, 2010.

\bibitem[Hou et~al.(2024)Hou, Wang, Pan, Wang, Xue, and Fu]{hou2024tamma}
Jiawei Hou, Tianyu Wang, Tongying Pan, Shouyan Wang, Xiangyang Xue, and Yanwei Fu.
\newblock Ta{MM}a: Target-driven multi-subscene mobile manipulation.
\newblock In \emph{8th Annual Conference on Robot Learning}, 2024.

\bibitem[Huang et~al.(2023{\natexlab{a}})Huang, Mees, Zeng, and Burgard]{huang2023visual}
Chenguang Huang, Oier Mees, Andy Zeng, and Wolfram Burgard.
\newblock Visual language maps for robot navigation.
\newblock In \emph{2023 IEEE International Conference on Robotics and Automation (ICRA)}, pages 10608--10615. IEEE, 2023{\natexlab{a}}.

\bibitem[Huang et~al.(2024)Huang, Li, Hui, and Yeung]{hhuang2024photoslam}
Huajian Huang, Longwei Li, Cheng Hui, and Sai-Kit Yeung.
\newblock Photo-slam: Real-time simultaneous localization and photorealistic mapping for monocular, stereo, and rgb-d cameras.
\newblock In \emph{Proceedings of the IEEE/CVF Conference on Computer Vision and Pattern Recognition}, 2024.

\bibitem[Huang et~al.(2022)Huang, Abbeel, Pathak, and Mordatch]{huang2022language}
Wenlong Huang, Pieter Abbeel, Deepak Pathak, and Igor Mordatch.
\newblock Language models as zero-shot planners: Extracting actionable knowledge for embodied agents.
\newblock In \emph{International conference on machine learning}, pages 9118--9147. PMLR, 2022.

\bibitem[Huang et~al.(2023{\natexlab{b}})Huang, Wang, Zhang, Li, Wu, and Fei-Fei]{huang2023voxposer}
Wenlong Huang, Chen Wang, Ruohan Zhang, Yunzhu Li, Jiajun Wu, and Li Fei-Fei.
\newblock Voxposer: Composable 3d value maps for robotic manipulation with language models.
\newblock \emph{arXiv preprint arXiv:2307.05973}, 2023{\natexlab{b}}.

\bibitem[Izadi et~al.(2011)Izadi, Kim, Hilliges, Molyneaux, Newcombe, Kohli, Shotton, Hodges, Freeman, Davison, et~al.]{izadi2011kinectfusion}
Shahram Izadi, David Kim, Otmar Hilliges, David Molyneaux, Richard Newcombe, Pushmeet Kohli, Jamie Shotton, Steve Hodges, Dustin Freeman, Andrew Davison, et~al.
\newblock Kinectfusion: real-time 3d reconstruction and interaction using a moving depth camera.
\newblock In \emph{Proceedings of the 24th annual ACM symposium on User interface software and technology}, pages 559--568, 2011.

\bibitem[Jiang et~al.(2024)Jiang, Lei, and Daniilidis]{jiang2024fisherrf}
Wen Jiang, Boshu Lei, and Kostas Daniilidis.
\newblock Fisherrf: Active view selection and mapping with radiance fields using fisher information.
\newblock In \emph{ECCV}, page 422–440, 2024.

\bibitem[Jin et~al.(2024)Jin, Gao, Wang, Wu, Lu, Xu, and Gao]{jin2024gs}
Rui Jin, Yuman Gao, Yingjian Wang, Yuze Wu, Haojian Lu, Chao Xu, and Fei Gao.
\newblock Gs-planner: A gaussian-splatting-based planning framework for active high-fidelity reconstruction.
\newblock In \emph{2024 IEEE/RSJ International Conference on Intelligent Robots and Systems (IROS)}, pages 11202--11209. IEEE, 2024.

\bibitem[Keetha et~al.(2023)Keetha, Karhade, Jatavallabhula, Yang, Scherer, Ramanan, and Luiten]{keetha2023splatam}
Nikhil Keetha, Jay Karhade, Krishna~Murthy Jatavallabhula, Gengshan Yang, Sebastian Scherer, Deva Ramanan, and Jonathon Luiten.
\newblock Splatam: Splat, track \& map 3d gaussians for dense rgb-d slam.
\newblock \emph{arXiv preprint arXiv:2312.02126}, 2023.

\bibitem[Kerbl et~al.(2023)Kerbl, Kopanas, Leimkuhler, and Drettakis]{kerbl3dgaussians}
Bernhard Kerbl, Georgios Kopanas, Thomas Leimkuhler, and George Drettakis.
\newblock 3d gaussian splatting for real-time radiance field rendering.
\newblock \emph{ACM Transactions on Graphics}, 42\penalty0 (4), 2023.

\bibitem[Khanna et~al.(2024)Khanna, Ramrakhya, Chhablani, Yenamandra, Gervet, Chang, Kira, Chaplot, Batra, and Mottaghi]{khanna2024goat}
Mukul Khanna, Ram Ramrakhya, Gunjan Chhablani, Sriram Yenamandra, Theophile Gervet, Matthew Chang, Zsolt Kira, Devendra~Singh Chaplot, Dhruv Batra, and Roozbeh Mottaghi.
\newblock Goat-bench: A benchmark for multi-modal lifelong navigation.
\newblock In \emph{Proceedings of the IEEE/CVF Conference on Computer Vision and Pattern Recognition}, pages 16373--16383, 2024.

\bibitem[Kim and Eustice(2015)]{kim2015active}
Ayoung Kim and Ryan~M Eustice.
\newblock Active visual slam for robotic area coverage: Theory and experiment.
\newblock \emph{The International Journal of Robotics Research}, 34\penalty0 (4-5):\penalty0 457--475, 2015.

\bibitem[Kuang et~al.(2024)Kuang, Yan, Zhao, Zhou, and Zha]{kuang2024active}
Zijia Kuang, Zike Yan, Hao Zhao, Guyue Zhou, and Hongbin Zha.
\newblock Active neural mapping at scale.
\newblock In \emph{2024 IEEE/RSJ International Conference on Intelligent Robots and Systems (IROS)}, pages 7152--7159. IEEE, 2024.

\bibitem[Kunstner et~al.(2019)Kunstner, Hennig, and Balles]{kunstner2019limitations}
Frederik Kunstner, Philipp Hennig, and Lukas Balles.
\newblock Limitations of the empirical fisher approximation for natural gradient descent.
\newblock \emph{Advances in neural information processing systems}, 32, 2019.

\bibitem[Lee et~al.(2022)Lee, Chen, Wang, Liniger, Kumar, and Yu]{lee2022uncertainty}
Soomin Lee, Le Chen, Jiahao Wang, Alexander Liniger, Suryansh Kumar, and Fisher Yu.
\newblock Uncertainty guided policy for active robotic 3d reconstruction using neural radiance fields.
\newblock \emph{IEEE Robotics and Automation Letters}, 7\penalty0 (4):\penalty0 12070--12077, 2022.

\bibitem[Liu et~al.(2024)Liu, Orru, Paxton, Shafiullah, and Pinto]{liu2024okrobot}
Peiqi Liu, Yaswanth Orru, Chris Paxton, Nur Muhammad~Mahi Shafiullah, and Lerrel Pinto.
\newblock Ok-robot: What really matters in integrating open-knowledge models for robotics.
\newblock \emph{arXiv preprint arXiv:2401.12202}, 2024.

\bibitem[Liu et~al.(2023)Liu, Prabhu, Cladera, Miller, Zhou, Taylor, and Kumar]{liu2023active}
Xu Liu, Ankit Prabhu, Fernando Cladera, Ian~D Miller, Lifeng Zhou, Camillo~J Taylor, and Vijay Kumar.
\newblock Active metric-semantic mapping by multiple aerial robots.
\newblock In \emph{2023 IEEE International Conference on Robotics and Automation (ICRA)}, pages 3282--3288. IEEE, 2023.

\bibitem[Lluvia et~al.(2021)Lluvia, Lazkano, and Ansuategi]{lluvia2021active-survey}
Iker Lluvia, Elena Lazkano, and Ander Ansuategi.
\newblock Active mapping and robot exploration: A survey.
\newblock \emph{Sensors}, 21\penalty0 (7):\penalty0 2445, 2021.

\bibitem[Long et~al.(2024{\natexlab{a}})Long, Cai, Wang, Zhan, and Dong]{long2024instructnav}
Yuxing Long, Wenzhe Cai, Hongcheng Wang, Guanqi Zhan, and Hao Dong.
\newblock Instructnav: Zero-shot system for generic instruction navigation in unexplored environment.
\newblock \emph{CoRL}, 2024{\natexlab{a}}.

\bibitem[Long et~al.(2024{\natexlab{b}})Long, Li, Cai, and Dong]{long2024discuss}
Yuxing Long, Xiaoqi Li, Wenzhe Cai, and Hao Dong.
\newblock Discuss before moving: Visual language navigation via multi-expert discussions.
\newblock In \emph{2024 IEEE International Conference on Robotics and Automation (ICRA)}, pages 17380--17387. IEEE, 2024{\natexlab{b}}.

\bibitem[MacKay(1992)]{bayesian-interpolation}
David J.~C. MacKay.
\newblock {Bayesian Interpolation}.
\newblock \emph{Neural Computation}, 4\penalty0 (3):\penalty0 415--447, 1992.

\bibitem[Martens(2020)]{martens2020new}
James Martens.
\newblock New insights and perspectives on the natural gradient method.
\newblock \emph{Journal of Machine Learning Research}, 21\penalty0 (146):\penalty0 1--76, 2020.

\bibitem[Matsuki et~al.(2024)Matsuki, Murai, Kelly, and Davison]{matsuki2023gaussian}
Hidenobu Matsuki, Riku Murai, Paul H.~J. Kelly, and Andrew~J. Davison.
\newblock {G}aussian {S}platting {SLAM}.
\newblock In \emph{Proceedings of the IEEE/CVF Conference on Computer Vision and Pattern Recognition}, 2024.

\bibitem[Mirjalili et~al.(2023)Mirjalili, Krawez, and Burgard]{mirjalili2023fmloc}
Reihaneh Mirjalili, Michael Krawez, and Wolfram Burgard.
\newblock Fm-loc: Using foundation models for improved vision-based localization.
\newblock In \emph{2023 IEEE/RSJ International Conference on Intelligent Robots and Systems (IROS)}, pages 1381--1387. IEEE, 2023.

\bibitem[Morilla-Cabello et~al.(2023)Morilla-Cabello, Mur-Labadia, Martinez-Cantin, and Montijano]{morilla2023robust}
David Morilla-Cabello, Lorenzo Mur-Labadia, Ruben Martinez-Cantin, and Eduardo Montijano.
\newblock Robust fusion for bayesian semantic mapping.
\newblock In \emph{2023 IEEE/RSJ International Conference on Intelligent Robots and Systems (IROS)}, pages 76--81. IEEE, 2023.

\bibitem[Mur-Artal and Tard{\'o}s(2017)]{mur2017orb}
Raul Mur-Artal and Juan~D Tard{\'o}s.
\newblock Orb-slam2: An open-source slam system for monocular, stereo, and rgb-d cameras.
\newblock \emph{IEEE transactions on robotics}, 33\penalty0 (5):\penalty0 1255--1262, 2017.

\bibitem[Mur-Artal et~al.(2015)Mur-Artal, Montiel, and Tardos]{mur2015orb}
Raul Mur-Artal, Jose Maria~Martinez Montiel, and Juan~D Tardos.
\newblock Orb-slam: a versatile and accurate monocular slam system.
\newblock \emph{IEEE transactions on robotics}, 31\penalty0 (5):\penalty0 1147--1163, 2015.

\bibitem[Newcombe et~al.(2011)Newcombe, Lovegrove, and Davison]{newcombe2011dtam}
Richard~A Newcombe, Steven~J Lovegrove, and Andrew~J Davison.
\newblock Dtam: Dense tracking and mapping in real-time.
\newblock In \emph{2011 international conference on computer vision}, pages 2320--2327. IEEE, 2011.

\bibitem[OpenAI(2024)]{gpt-4o}
OpenAI.
\newblock Gpt-4o system card, 2024.
\newblock https://openai.com/index/gpt-4o-system-card/.

\bibitem[Pan et~al.(2022)Pan, Lai, Song, and Huang]{pan2022activenerf}
Xuran Pan, Zihang Lai, Shiji Song, and Gao Huang.
\newblock Activenerf: Learning where to see with uncertainty estimation.
\newblock In \emph{ECCV}, pages 230--246. Springer, 2022.

\bibitem[P{\'a}zman(1986)]{pazman1986foundations}
Andrej P{\'a}zman.
\newblock \emph{Foundations of optimum experimental design}.
\newblock Springer Dordrecht, 1986.

\bibitem[Placed et~al.(2022)Placed, Rodr{\'\i}guez, Tard{\'o}s, and Castellanos]{placed2022explorb}
Julio~A Placed, Juan J~G{\'o}mez Rodr{\'\i}guez, Juan~D Tard{\'o}s, and Jos{\'e}~A Castellanos.
\newblock Explorb-slam: Active visual slam exploiting the pose-graph topology.
\newblock In \emph{Iberian Robotics conference}, pages 199--210. Springer, 2022.

\bibitem[Placed et~al.(2023)Placed, Strader, Carrillo, Atanasov, Indelman, Carlone, and Castellanos]{placed2023survey}
Julio~A Placed, Jared Strader, Henry Carrillo, Nikolay Atanasov, Vadim Indelman, Luca Carlone, and Jos{\'e}~A Castellanos.
\newblock A survey on active simultaneous localization and mapping: State of the art and new frontiers.
\newblock \emph{IEEE Transactions on Robotics}, 2023.

\bibitem[Qin et~al.(2024)Qin, Li, Zhou, Wang, and Pfister]{qin2024langsplat}
Minghan Qin, Wanhua Li, Jiawei Zhou, Haoqian Wang, and Hanspeter Pfister.
\newblock Langsplat: 3d language gaussian splatting.
\newblock In \emph{Proceedings of the IEEE/CVF Conference on Computer Vision and Pattern Recognition}, pages 20051--20060, 2024.

\bibitem[Ramakrishnan et~al.(2021)Ramakrishnan, Gokaslan, Wijmans, Maksymets, Clegg, Turner, Undersander, Galuba, Westbury, Chang, Savva, Zhao, and Batra]{ramakrishnan2021hm3d}
Santhosh~Kumar Ramakrishnan, Aaron Gokaslan, Erik Wijmans, Oleksandr Maksymets, Alexander Clegg, John~M Turner, Eric Undersander, Wojciech Galuba, Andrew Westbury, Angel~X Chang, Manolis Savva, Yili Zhao, and Dhruv Batra.
\newblock Habitat-matterport 3d dataset ({HM}3d): 1000 large-scale 3d environments for embodied {AI}.
\newblock In \emph{Thirty-fifth Conference on Neural Information Processing Systems Datasets and Benchmarks Track}, 2021.

\bibitem[Rana et~al.(2023)Rana, Haviland, Garg, Abou-Chakra, Reid, and Suenderhauf]{rana2023sayplan}
Krishan Rana, Jesse Haviland, Sourav Garg, Jad Abou-Chakra, Ian~D Reid, and Niko Suenderhauf.
\newblock Sayplan: Grounding large language models using 3d scene graphs for scalable task planning.
\newblock \emph{CoRR}, 2023.

\bibitem[Rodr{\'\i}guez-Ar{\'e}valo et~al.(2018)Rodr{\'\i}guez-Ar{\'e}valo, Neira, and Castellanos]{rodriguez2018importance}
Mar{\'\i}a~L Rodr{\'\i}guez-Ar{\'e}valo, Jos{\'e} Neira, and Jos{\'e}~A Castellanos.
\newblock On the importance of uncertainty representation in active slam.
\newblock \emph{IEEE Transactions on Robotics}, 34\penalty0 (3):\penalty0 829--834, 2018.

\bibitem[Rückin et~al.(2023)Rückin, Magistri, Stachniss, and Popović]{informative-uav}
Julius Rückin, Federico Magistri, Cyrill Stachniss, and Marija Popović.
\newblock An informative path planning framework for active learning in uav-based semantic mapping.
\newblock \emph{IEEE Transactions on Robotics}, 39\penalty0 (6):\penalty0 4279--4296, 2023.

\bibitem[S et~al.(2024)S, Melnik, and Nandi]{s2024cognitive}
Arjun~P S, Andrew Melnik, and Gora~Chand Nandi.
\newblock Cognitive planning for object goal navigation using generative {AI} models.
\newblock In \emph{NeurIPS 2024 Workshop on Open-World Agents}, 2024.

\bibitem[Schervish(2012)]{schervish2012theory}
M.J. Schervish.
\newblock \emph{Theory of Statistics}.
\newblock Springer New York, 2012.

\bibitem[Sermanet et~al.(2024)Sermanet, Ding, Zhao, Xia, Dwibedi, Gopalakrishnan, Chan, Dulac-Arnold, Maddineni, Joshi, et~al.]{sermanet2024robovqa}
Pierre Sermanet, Tianli Ding, Jeffrey Zhao, Fei Xia, Debidatta Dwibedi, Keerthana Gopalakrishnan, Christine Chan, Gabriel Dulac-Arnold, Sharath Maddineni, Nikhil~J Joshi, et~al.
\newblock Robovqa: Multimodal long-horizon reasoning for robotics.
\newblock In \emph{2024 IEEE International Conference on Robotics and Automation (ICRA)}, pages 645--652. IEEE, 2024.

\bibitem[Shafiullah et~al.(2023)Shafiullah, Paxton, Pinto, Chintala, and Szlam]{shafiullah2023clipfields}
Nur Muhammad~Mahi Shafiullah, Chris Paxton, Lerrel Pinto, Soumith Chintala, and Arthur Szlam.
\newblock {CLIP}-fields: Weakly supervised semantic fields for robotic memory.
\newblock In \emph{ICRA2023 Workshop on Pretraining for Robotics (PT4R)}, 2023.

\bibitem[Shah et~al.(2023)Shah, Equi, Osi{\'n}ski, Xia, Ichter, and Levine]{shah2023navigation}
Dhruv Shah, Michael~Robert Equi, B{\l}a{\.z}ej Osi{\'n}ski, Fei Xia, Brian Ichter, and Sergey Levine.
\newblock Navigation with large language models: Semantic guesswork as a heuristic for planning.
\newblock In \emph{Conference on Robot Learning}, pages 2683--2699. PMLR, 2023.

\bibitem[Shannon(1948)]{shannon1948mathematical}
Claude~Elwood Shannon.
\newblock A mathematical theory of communication.
\newblock \emph{The Bell system technical journal}, 27\penalty0 (3):\penalty0 379--423, 1948.

\bibitem[Shen et~al.(2021)Shen, Ruiz, Agudo, and Moreno{-}Noguer]{shen2021snerf}
Jianxiong Shen, Adria Ruiz, Antonio Agudo, and Francesc Moreno{-}Noguer.
\newblock Stochastic neural radiance fields: Quantifying uncertainty in implicit 3d representations.
\newblock \emph{CoRR}, abs/2109.02123, 2021.

\bibitem[Shen et~al.(2022)Shen, Agudo, Moreno-Noguer, and Ruiz]{cf-nerf}
Jianxiong Shen, Antonio Agudo, Francesc Moreno-Noguer, and Adria Ruiz.
\newblock Conditional-flow nerf: Accurate 3d modelling with reliable uncertainty quantification.
\newblock In \emph{ECCV}, 2022.

\bibitem[Shi et~al.(2024)Shi, Wang, Duan, and Guan]{shi2024language}
Jin-Chuan Shi, Miao Wang, Hao-Bin Duan, and Shao-Hua Guan.
\newblock Language embedded 3d gaussians for open-vocabulary scene understanding.
\newblock In \emph{Proceedings of the IEEE/CVF Conference on Computer Vision and Pattern Recognition}, pages 5333--5343, 2024.

\bibitem[Song et~al.(2023)Song, Wu, Washington, Sadler, Chao, and Su]{song2023llm}
Chan~Hee Song, Jiaman Wu, Clayton Washington, Brian~M Sadler, Wei-Lun Chao, and Yu Su.
\newblock Llm-planner: Few-shot grounded planning for embodied agents with large language models.
\newblock In \emph{Proceedings of the IEEE/CVF International Conference on Computer Vision}, pages 2998--3009, 2023.

\bibitem[Stachniss et~al.(2004)Stachniss, Hahnel, and Burgard]{stachniss2004exploration}
Cyrill Stachniss, Dirk Hahnel, and Wolfram Burgard.
\newblock Exploration with active loop-closing for fastslam.
\newblock In \emph{2004 IEEE/RSJ International Conference on Intelligent Robots and Systems (IROS)(IEEE Cat. No. 04CH37566)}, pages 1505--1510. IEEE, 2004.

\bibitem[Sun et~al.(2020)Sun, Wu, Xu, Sarma, Yang, and Kong]{sun2020frontier}
Zezhou Sun, Banghe Wu, Cheng-Zhong Xu, Sanjay~E Sarma, Jian Yang, and Hui Kong.
\newblock Frontier detection and reachability analysis for efficient 2d graph-slam based active exploration.
\newblock In \emph{2020 IEEE/RSJ International Conference on Intelligent Robots and Systems (IROS)}, pages 2051--2058. IEEE, 2020.

\bibitem[S{\"u}nderhauf et~al.(2023)S{\"u}nderhauf, Abou-Chakra, and Miller]{sunderhauf2023density}
Niko S{\"u}nderhauf, Jad Abou-Chakra, and Dimity Miller.
\newblock Density-aware nerf ensembles: Quantifying predictive uncertainty in neural radiance fields.
\newblock In \emph{2023 IEEE International Conference on Robotics and Automation (ICRA)}, pages 9370--9376. IEEE, 2023.

\bibitem[Szot et~al.(2021)Szot, Clegg, Undersander, Wijmans, Zhao, Turner, Maestre, Mukadam, Chaplot, Maksymets, et~al.]{szot2021habitat}
Andrew Szot, Alexander Clegg, Eric Undersander, Erik Wijmans, Yili Zhao, John Turner, Noah Maestre, Mustafa Mukadam, Devendra~Singh Chaplot, Oleksandr Maksymets, et~al.
\newblock Habitat 2.0: Training home assistants to rearrange their habitat.
\newblock \emph{Advances in neural information processing systems}, 34:\penalty0 251--266, 2021.

\bibitem[Tao et~al.(2024{\natexlab{a}})Tao, Liu, Spasojevic, Agarwal, and Kumar]{tao20243d}
Yuezhan Tao, Xu Liu, Igor Spasojevic, Saurav Agarwal, and Vijay Kumar.
\newblock 3d active metric-semantic slam.
\newblock \emph{IEEE Robotics and Automation Letters}, 9\penalty0 (3):\penalty0 2989--2996, 2024{\natexlab{a}}.

\bibitem[Tao et~al.(2024{\natexlab{b}})Tao, Ong, Murali, Spasojevic, Chaudhari, and Kumar]{tao2024rt}
Yuezhan Tao, Dexter Ong, Varun Murali, Igor Spasojevic, Pratik Chaudhari, and Vijay Kumar.
\newblock Rt-guide: Real-time gaussian splatting for information-driven exploration.
\newblock \emph{arXiv preprint arXiv:2409.18122}, 2024{\natexlab{b}}.

\bibitem[Team et~al.(2024)Team, Ghosh, Walke, Pertsch, Black, Mees, Dasari, Hejna, Kreiman, Xu, et~al.]{team2024octo}
Octo~Model Team, Dibya Ghosh, Homer Walke, Karl Pertsch, Kevin Black, Oier Mees, Sudeep Dasari, Joey Hejna, Tobias Kreiman, Charles Xu, et~al.
\newblock Octo: An open-source generalist robot policy.
\newblock \emph{arXiv preprint arXiv:2405.12213}, 2024.

\bibitem[Teed and Deng(2021)]{teed2021droid}
Zachary Teed and Jia Deng.
\newblock Droid-slam: Deep visual slam for monocular, stereo, and rgb-d cameras.
\newblock \emph{Advances in neural information processing systems}, 34:\penalty0 16558--16569, 2021.

\bibitem[Triantafyllidis et~al.(2024)Triantafyllidis, Christianos, and Li]{triantafyllidis2024intrinsic}
Eleftherios Triantafyllidis, Filippos Christianos, and Zhibin Li.
\newblock Intrinsic language-guided exploration for complex long-horizon robotic manipulation tasks.
\newblock In \emph{2024 IEEE International Conference on Robotics and Automation (ICRA)}, pages 7493--7500. IEEE, 2024.

\bibitem[Trivun et~al.(2015)Trivun, {\v{S}}alaka, Osmankovi{\'c}, Velagi{\'c}, and Osmi{\'c}]{trivun2015active}
Darko Trivun, Edin {\v{S}}alaka, Dinko Osmankovi{\'c}, Jasmin Velagi{\'c}, and Nedim Osmi{\'c}.
\newblock Active slam-based algorithm for autonomous exploration with mobile robot.
\newblock In \emph{2015 IEEE International Conference on Industrial Technology (ICIT)}, pages 74--79. IEEE, 2015.

\bibitem[Vidal-Calleja et~al.(2006)Vidal-Calleja, Davison, Andrade-Cetto, and Murray]{vidal2006active}
Teresa Vidal-Calleja, Andrew~J Davison, Juan Andrade-Cetto, and David~William Murray.
\newblock Active control for single camera slam.
\newblock In \emph{Proceedings 2006 IEEE International Conference on Robotics and Automation, 2006. ICRA 2006.}, pages 1930--1936. IEEE, 2006.

\bibitem[Wang et~al.(2004)Wang, Bovik, Sheikh, and Simoncelli]{wang2004image}
Zhou Wang, Alan~C Bovik, Hamid~R Sheikh, and Eero~P Simoncelli.
\newblock Image quality assessment: from error visibility to structural similarity.
\newblock \emph{IEEE transactions on image processing}, 13\penalty0 (4):\penalty0 600--612, 2004.

\bibitem[Wei et~al.(2022)Wei, Wang, Schuurmans, Bosma, Xia, Chi, Le, Zhou, et~al.]{wei2022chain}
Jason Wei, Xuezhi Wang, Dale Schuurmans, Maarten Bosma, Fei Xia, Ed Chi, Quoc~V Le, Denny Zhou, et~al.
\newblock Chain-of-thought prompting elicits reasoning in large language models.
\newblock \emph{Advances in neural information processing systems}, 35:\penalty0 24824--24837, 2022.

\bibitem[Xia et~al.(2018)Xia, R.~Zamir, He, Sax, Malik, and Savarese]{xiazamirhe2018gibsonenv}
Fei Xia, Amir R.~Zamir, Zhi-Yang He, Alexander Sax, Jitendra Malik, and Silvio Savarese.
\newblock Gibson {Env}: real-world perception for embodied agents.
\newblock In \emph{Computer Vision and Pattern Recognition (CVPR), 2018 IEEE Conference on}. IEEE, 2018.

\bibitem[Yamauchi(1997)]{yamauchi1997frontier}
Brian Yamauchi.
\newblock A frontier-based approach for autonomous exploration.
\newblock In \emph{Proceedings 1997 IEEE International Symposium on Computational Intelligence in Robotics and Automation CIRA'97.'Towards New Computational Principles for Robotics and Automation'}, pages 146--151. IEEE, 1997.

\bibitem[Yan et~al.(2023{\natexlab{a}})Yan, Qu, Wang, Xu, Wang, Zhao, and Li]{yan2023gs}
Chi Yan, Delin Qu, Dong Wang, Dan Xu, Zhigang Wang, Bin Zhao, and Xuelong Li.
\newblock Gs-slam: Dense visual slam with 3d gaussian splatting.
\newblock \emph{arXiv preprint arXiv:2311.11700}, 2023{\natexlab{a}}.

\bibitem[Yan et~al.(2023{\natexlab{b}})Yan, Liu, Quan, Chen, and Fu]{yan2023activeIO}
Dongyu Yan, Jianheng Liu, Fengyu Quan, Haoyao Chen, and Mengmeng Fu.
\newblock Active implicit object reconstruction using uncertainty-guided next-best-view optimization.
\newblock \emph{IEEE Robotics and Automation Letters}, 2023{\natexlab{b}}.

\bibitem[Yan et~al.(2023{\natexlab{c}})Yan, Yang, and Zha]{yan2023active-neural-mapping}
Zike Yan, Haoxiang Yang, and Hongbin Zha.
\newblock Active neural mapping.
\newblock In \emph{ICCV}, 2023{\natexlab{c}}.

\bibitem[Yokoyama et~al.(2024)Yokoyama, Ha, Batra, Wang, and Bucher]{yokoyama2024vlfm}
Naoki Yokoyama, Sehoon Ha, Dhruv Batra, Jiuguang Wang, and Bernadette Bucher.
\newblock Vlfm: Vision-language frontier maps for zero-shot semantic navigation.
\newblock In \emph{2024 IEEE International Conference on Robotics and Automation (ICRA)}, pages 42--48. IEEE, 2024.

\bibitem[Yu et~al.(2024)Yu, Hari, Srinivas, El-Refai, Rashid, Kim, Kerr, Cheng, Irshad, Balakrishna, et~al.]{yulanguage}
Justin Yu, Kush Hari, Kishore Srinivas, Karim El-Refai, Adam Rashid, Chung~Min Kim, Justin Kerr, Richard Cheng, Muhammad~Zubair Irshad, Ashwin Balakrishna, et~al.
\newblock Language-embedded gaussian splats (legs): Incrementally building room-scale representations with a mobile robot.
\newblock In \emph{IROS}, 2024.

\bibitem[Zhan et~al.(2022)Zhan, Zheng, Xu, Reid, and Rezatofighi]{zhan2022activermap}
Huangying Zhan, Jiyang Zheng, Yi Xu, Ian Reid, and Hamid Rezatofighi.
\newblock Activermap: Radiance field for active mapping and planning, 2022.

\bibitem[Zhang et~al.(2018)Zhang, Isola, Efros, Shechtman, and Wang]{zhang2018unreasonable}
Richard Zhang, Phillip Isola, Alexei~A Efros, Eli Shechtman, and Oliver Wang.
\newblock The unreasonable effectiveness of deep features as a perceptual metric.
\newblock In \emph{CVPR}, pages 586--595, 2018.

\bibitem[Zhou et~al.(2024{\natexlab{a}})Zhou, Hong, and Wu]{Zhou_Hong_Wu_2024}
Gengze Zhou, Yicong Hong, and Qi Wu.
\newblock Navgpt: Explicit reasoning in vision-and-language navigation with large language models.
\newblock \emph{Proceedings of the AAAI Conference on Artificial Intelligence}, 38\penalty0 (7):\penalty0 7641--7649, 2024{\natexlab{a}}.

\bibitem[Zhou et~al.(2024{\natexlab{b}})Zhou, Chang, Jiang, Fan, Zhu, Xu, Chari, You, Wang, and Kadambi]{zhou2024feature}
Shijie Zhou, Haoran Chang, Sicheng Jiang, Zhiwen Fan, Zehao Zhu, Dejia Xu, Pradyumna Chari, Suya You, Zhangyang Wang, and Achuta Kadambi.
\newblock Feature 3dgs: Supercharging 3d gaussian splatting to enable distilled feature fields.
\newblock In \emph{Proceedings of the IEEE/CVF Conference on Computer Vision and Pattern Recognition}, pages 21676--21685, 2024{\natexlab{b}}.

\end{thebibliography}
}

\clearpage

\section{Appendix}
In this supplementary material, we discuss things we left over in our main paper due to page constraints. We provide more background information about 3DGS and the derivative with respect to localization parameters in Sec.~\ref{sec:supp-background}. We discussed more implementation details in Sec.~\ref{sec:supp-impl} and provided the prompt we use for the Multimodal LLM and an example of the interaction in Sec.~\ref{sec:supp-prompt} to help reproduce our results. {\bf The source code of this project will be made public soon.}
We also include detailed qualitative results in Sec.~\ref{sec:supp-qual}.

\subsection{Additional Background Information}~\label{sec:supp-background}
For the completeness of our method, we also provide the key definition for the 3D Gaussian Splatting backbone~\cite{kerbl3dgaussians}  and 3D Gaussian SLAM~\cite{matsuki2023gaussian}.
In 3D Gaussian Splatting~\cite{kerbl3dgaussians}, the rendered pixel color is calculated by composing all 3D Gaussians projected in a tile.
\begin{equation}
\label{eq:agslam-supp-render}
\hat{C}(\text{r}) = \sum_{i=1}^{N_s} T_i \left(1-\exp(-\sigma_i \delta_i) \right) \bc_i
\end{equation}
\begin{equation}
\alpha_i=\text{exp}\left(-\sum_{j=1}^{i-1}\sigma_j\delta_j\right)(1-\text{exp}(-\sigma_i\delta_i))    
\end{equation}

$\delta_i=t_{i+1}-t_i$ represents the distance between adjacent samples, and $N_s$ indicates the number of samples. $\bc_i$ is the color of each 3D Gaussian given the current view direction ${\bf d}$ and $\sigma_i$ is given by evaluating a 2D Gaussian with covariance $\Sigma$.

The Jacobian of the localization parameters are defined as:

\begin{align}
    \mpd{\meanC}{\camCW} =  \begin{bmatrix}\identity &-\meanC^\times\end{bmatrix}
    \text{ and }
    \mpd{\matW}{\camCW} = \begin{bmatrix}
    \mathbf{0} & -\matW_{:, 1}^\times \\
    \mathbf{0} & -\matW_{:, 2}^\times \\
    \mathbf{0} & -\matW_{:, 3}^\times\\
    \end{bmatrix}\label{eqn:grad_meanC_camCW_W_camCW}
    ~, 
\end{align}
where ${}^{\times}$ denotes the skew symmetric matrix of a 3D vector, and $\matW_{:, i}$ refers to the $i$th column of the matrix.

Unlike PSNR, the EIG can be computed without ground truth images, making it possible to perform view selection during exploration. 

\subsection{Implementation Details}\label{sec:supp-impl}

The 2D occupancy map's resolution is 5cm. For each single frontier pixel on the 2D map, we add 200 3D Gaussians, which are uniformly distributed in the 3D cube above each frontier pixel. Other parameters like color, opacity, and scale are generated uniformly between 0 and 1. When there are frontiers on the 2D map, we choose the next frontier by querying LLM as stated in the main paper. When no frontier exists, we select the top 20\% of Gaussians with the highest score. These Gaussians are grouped using DBSCAN~\cite{ester1996density}. The largest cluster is selected for candidate pose generation. Candidates are uniformly sampled in the range between 0.3m to 1m, facing towards the selected position. Only the poses in free space are kept for path-level selection.
The importance factor $\eta$ in Eq.~\ref{eq:agslam-path} is set to 5 across all experiments.

We compute the Expected Information Gain (EIG) for each global candidate and use A* to plan a path to each of them. In order to prevent a twisted path, we consider locations 0.15m (3 pixels) away from the current robot position as neighbors and set the robot width to 3 pixels for collision check. However, the path planned by A* might have redundant waypoints, causing unnecessary turns for the robot. Therefore, we smooth the path by finding shortcuts. Specifically, for each waypoint $w_i$, if the path between waypoint $w_{i+2}$ and $w_i$ is collision-free, then we remove the intermediate waypoint $w_{i+1}$ from the path. Finally, we use a greedy follower for motion planning. If the angle between the heading direction of the robot and the relative next waypoint is larger than 5$\degree$, then we turn left or right to decrease the angle. Otherwise, we choose the forward action to approach the next waypoint. In such a way, we get a sequence of actions $\{a_i\}_{i=1}^{T}$ for each path. 

Given a sequence of actions $\{a_i\}_{i=1}^{T}$ for each path, we use forward dynamics to compute the future camera poses $\{c_i\}_{i=1}^{T}$. Initially, we use an intermediate variable $\bH''_{\text{obs}} \triangleq \bH''[\bw^*]$ to help compute expected information gain along the path. For each camera pose $x_i$, we compute its pose Hessian $\bH''_{\text{pose}}$ and the current model Hessian matrix $\bH''_{\text{cur}} \triangleq \bH''[\by | x_i, \bw^*]$. $\bH''_{\text{cur}}$ is then accumulated, and we update $\bH''_{\text{obs}}$ to evaluate the remaining poses on the path. We select the path that minimizes the objective given by Eq.~\ref{eq:agslam-path} for execution.

\subsection{Scenes Used for Evaluation}
\label{sec:supp-results}
Following previous literature~\cite{yan2023active-neural-mapping}, we use the following scenes for Gibson Dataset: \texttt{Greigsville, Denmark, Cantwell, Eudora, Pablo, Ribera, Swormville, Eastville, Elmira}. For HM3D we use the following scenes: \texttt{DBjEcHFg4oq, mscxX4KEBcB, QKGMrurUVbk, oPj9qMxrDEa, CETmJJqkhcK}.

\subsection{Detailed Version of Qualitative Results}\label{sec:supp-qual}

\begin{figure*}[t]
\setlength{\tabcolsep}{0pt}
	\centering
\begin{tabular}{ccc}
UPEN~\cite{upen} & FBE~\cite{yamauchi1997frontier} & Ours \\ 
\includegraphics[width=0.32\textwidth]{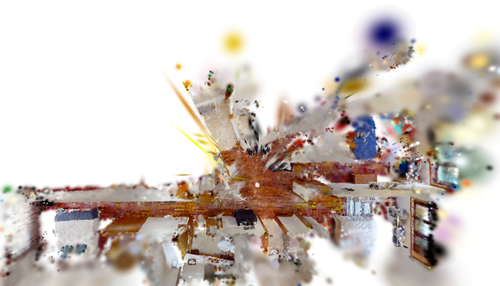} &
\includegraphics[width=0.32\textwidth]{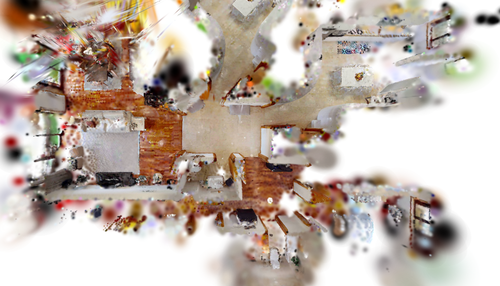} &
\includegraphics[width=0.32\textwidth]{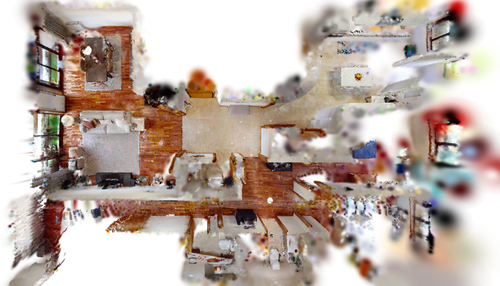} \\
\includegraphics[width=0.32\textwidth]{figures/hm3d-qual/oPj9qMxrDEa-upen-cropped.png} &
\includegraphics[width=0.32\textwidth]{figures/hm3d-qual/oPj9qMxrDEa-frontier-cropped.png} &
\includegraphics[width=0.32\textwidth]{figures/hm3d-qual/oPj9qMxrDEa-ours-cropped.png} \\
\includegraphics[width=0.32\textwidth]{figures/hm3d-qual/QKGMrurUVbk-upen-cropped.png} &
\includegraphics[width=0.32\textwidth]{figures/hm3d-qual/QKGMrurUVbk-frontier-cropped.png} &
\includegraphics[width=0.32\textwidth]{figures/hm3d-qual/QKGMrurUVbk-ours-cropped.png} \\
\end{tabular}
\caption{{\bf Qualitative Comparison for Final Scene Reconstruction on Habitat-Matterport 3D Dataset} mscxX4KEBcB~(top), oPj9qMxrDEa~(middle) and QKGMrurUVbk~(bottom) scenes. We provide top-down rendering for different methods. }\label{fig:qual-hm3d-large}
\end{figure*}

We provide larger versions of the qualitative rendering comparisons from the main paper, \cref{fig:qual-gibson-large} shows the Gibson scenes and \cref{fig:qual-hm3d-large} shows the HM3D scenes.

We also present qualitative comparisons on testing views from the Gibson dataset in \cref{fig:qual-render-gibson} and HM3D in \cref{fig:qual-render}.

\subsection{Example of Using Multimodal LLM}\label{sec:supp-prompt}

\begin{figure*}[t]
    \centering
    \includegraphics[width=0.86\linewidth]{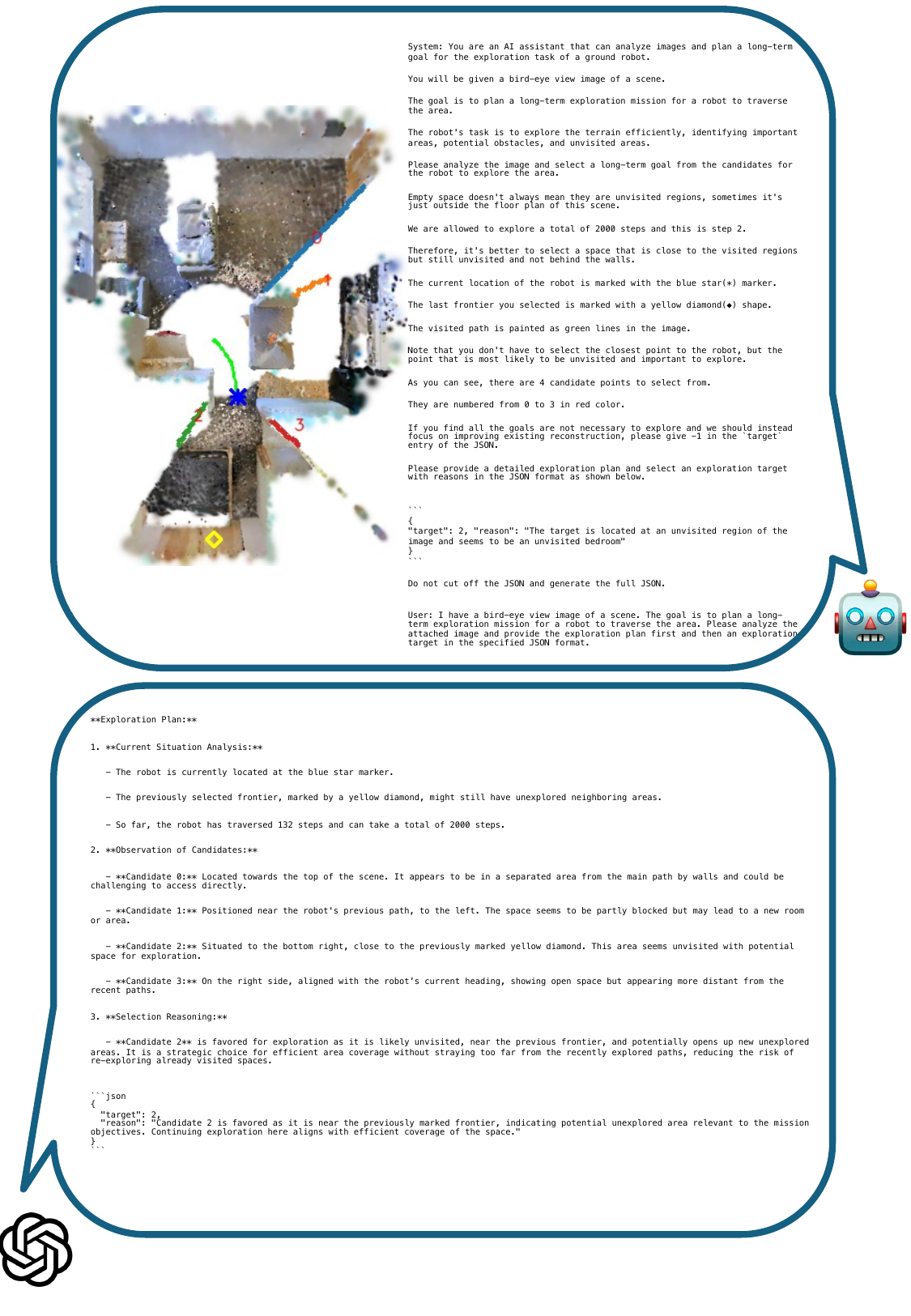}
    \caption{{\bf An Example about a Dialog with the Multi-modal LLM.} The chat box on the top is the question from our method, and the bottom is the response from GPT-4o, which we parsed to get a long-term goal for our path proposal. We provided the contextual information in both the textual prompt and the image.}
    \label{fig:qa-example}
\end{figure*}

We provide an example of our interaction with the multimodal LLM in \cref{fig:qa-example} and additionally provide our full text prompt to the LLM below.

\definecolor{codebackground}{rgb}{0.95,0.95,0.95}  %
\definecolor{codegray}{rgb}{0.5,0.5,0.5}          %
\definecolor{codeblue}{rgb}{0.13,0.13,1}          %
\definecolor{codepurple}{rgb}{0.58,0,0.82}        %

\lstset{
    breaklines=true,                 %
    breakatwhitespace=true,         %
    columns=fullflexible,           %
    frame=single,                   %
    numbers=left,                   %
    numberstyle=\tiny\color{codegray},  %
    xleftmargin=2em,               %
    keepspaces=true,               %
    showstringspaces=false,        %
    captionpos=b,                  %
    linewidth=\columnwidth,        %
    upquote=true,                  %
    tabsize=2,                     %
    backgroundcolor=\color{codebackground},
    basicstyle=\footnotesize\fontfamily{pcr}\selectfont,      %
    commentstyle=\color{codegray}\fontfamily{pcr}\selectfont, %
    keywordstyle=\color{codeblue}\fontfamily{pcr}\selectfont, %
    stringstyle=\color{codepurple}\fontfamily{pcr}\selectfont,%
    framexleftmargin=1mm,          %
    framexrightmargin=0mm,         %
    framesep=2mm,                  %
    rulesep=1mm,                   %
    framerule=0.4pt,              %
}

\begin{lstlisting}

<system> You are an AI assistant that can analyze images and plan a long-term goal for the exploration task of a ground robot.
You will be given a bird-eye view image of a scene.
The goal is to plan a long-term exploration mission for a robot to traverse the area.
The robot's task is to explore the terrain efficiently, identifying important areas, potential obstacles, and unvisited areas.
Please analyze the image and select a long-term goal from the candidates for the robot to explore the area.
Empty space doesn't always mean they are unvisited regions, sometimes it's just outside the floor plan of this scene.
We are allowed to explore a total of <TOTAL\_STEPS> steps and this is step <STEP\_ID>.
Therefore, it's better to select a space that is close to the visited regions but still unvisited and not behind the walls.
The current location of the robot is marked with the blue star(*) marker.
The last frontier you selected is marked with a yellow diamond shape. 
The visited path is painted as green lines in the image.
Note that you don't have to select the closest point to the robot, but the point that is most likely to be unvisited and important to explore.
As you can see, there are <NUM\_FRAME> candidate points to select from.
They are numbered from 0 to <NUM\_FRAME - 1> in red color.
If you find all the goals are not necessary to explore and we should instead focus on improving existing reconstruction, please give -1 in the `target` entry of the JSON.
Please provide a detailed exploration plan and select an exploration target with reasons in the JSON format as shown below.


```
{
"target": 2, "reason": "The target is located at an unvisited region of the image and seems to be an unvisited bedroom"
}
```
Do not cut off the JSON and generate the full JSON.
</system>


<user>: I have a bird-eye view image of a scene. The goal is to plan a long-term exploration mission for a robot to traverse the area. Please analyze the attached image and provide the exploration plan first and then an exploration target in the specified JSON format.
</user>

\end{lstlisting}

\begin{figure*}[p!]
\setlength{\tabcolsep}{0pt}
	\centering
\begin{tabular}{ccc}
ANS~\cite{chaplot2020learning} & Active-INR~\cite{yan2023active-neural-mapping} & UPEN~\cite{upen} \\
\includegraphics[width=0.31\textwidth,trim={2.5cm 2.5cm 2.5cm 2.5cm},clip]{figures/Qualitative/Greigsville_ANS_5_est.png} &
\includegraphics[width=0.31\textwidth,trim={2.5cm 2.5cm 2.5cm 2.5cm},clip]{figures/Qualitative/Greigsville_INR_gt.png} &
\includegraphics[width=0.31\textwidth,trim={2.5cm 2.5cm 2.5cm 2.5cm},clip]{figures/Qualitative/Greigsville_UPEN_gtpose.png} \\
ExplORB~\cite{placed2022explorb} & FBE~\cite{yamauchi1997frontier} & Ours \\ 
\includegraphics[width=0.31\textwidth,trim={2.5cm 2.5cm 2.5cm 2.5cm},clip]{figures/Qualitative/Greigsville_explORB.png} &
\includegraphics[width=0.31\textwidth]{figures/Qualitative/Greigsville_Frontier.png} &
\includegraphics[width=0.31\textwidth]{figures/Qualitative/Greigsville_ours.png} 
\\
ANS~\cite{chaplot2020learning} & Active-INR~\cite{yan2023active-neural-mapping} & UPEN~\cite{upen} 
\\
\includegraphics[width=0.31\textwidth,trim={0cm 5cm 0cm 0cm},clip]{figures/Qualitative/Ribera_ANS_5_est.png} &
\includegraphics[width=0.31\textwidth,trim={0cm 5cm 0cm 0cm},clip]{figures/Qualitative/Ribera_INR_gt.png} &
\includegraphics[width=0.31\textwidth,trim={0cm 5cm 0cm 0cm},clip]{figures/Qualitative/Ribera_UPEN_gtpose.png} \\
ExplORB~\cite{placed2022explorb} & FBE~\cite{yamauchi1997frontier} &  Ours \\ 
\includegraphics[width=0.31\textwidth,trim={0cm 3.5cm 0cm 0cm},clip]{figures/Qualitative/Ribera_explORB.png} &
\includegraphics[width=0.31\textwidth,trim={0cm 3.5cm 0cm 0cm},clip]{figures/Qualitative/Ribera_Frontier.png} &
\includegraphics[width=0.31\textwidth,trim={0cm 3.5cm 0cm 0cm},clip]{figures/Qualitative/Ribera_ours.png} 
\end{tabular}
\caption{{\bf Qualitative Comparison for Final Scene Reconstruction on Gibson Dataset} Greigsville~(top) and Ribera~(bottom) scenes. We provide top-down rendering for different methods. Note that UPEN and Active-INR use GT pose in this visualization.}\label{fig:qual-gibson-large}
\end{figure*}

\newcommand{\rendersize}{1.4in}

\begin{figure*}[!t]
\centering

\begin{tabular}{>{\begin{sideways}}p{0.4in}<{\end{sideways}}p{0.25\textwidth}p{0.25\textwidth}p{0.25\textwidth}}
 \qquad \qquad  ANS & \includegraphics[height=\rendersize]{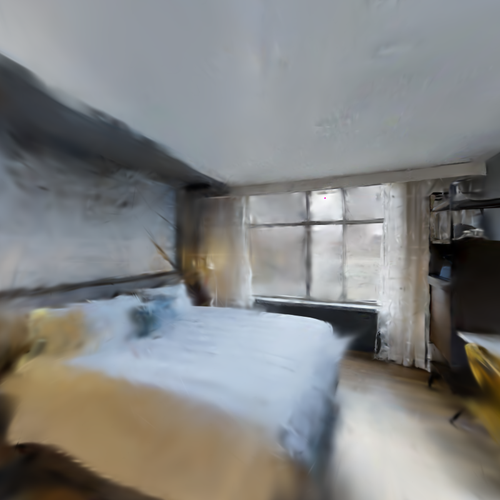} & \includegraphics[height=\rendersize]{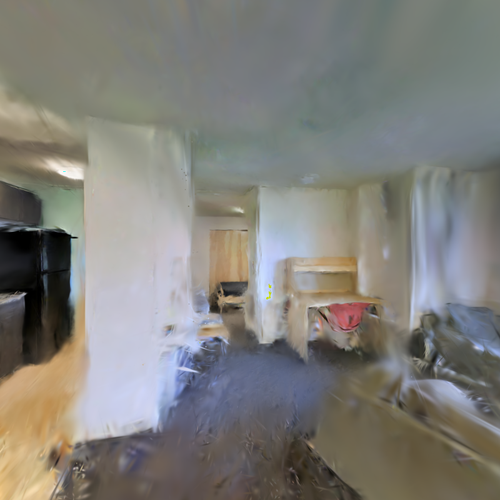} & \includegraphics[height=\rendersize]{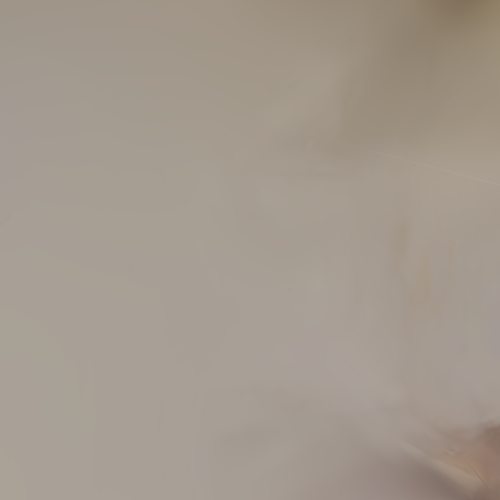} \\ 
 \qquad Active-INR (gt) & \includegraphics[height=\rendersize]{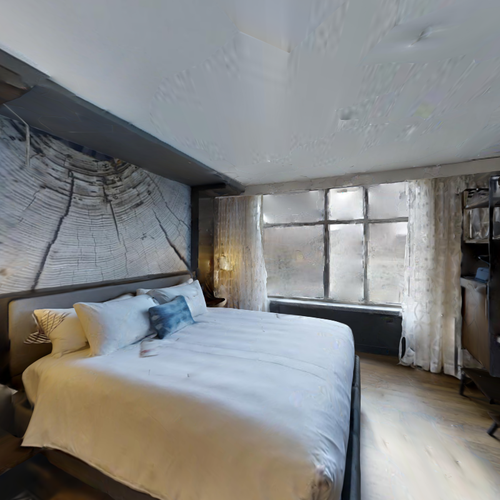} & \includegraphics[height=\rendersize]{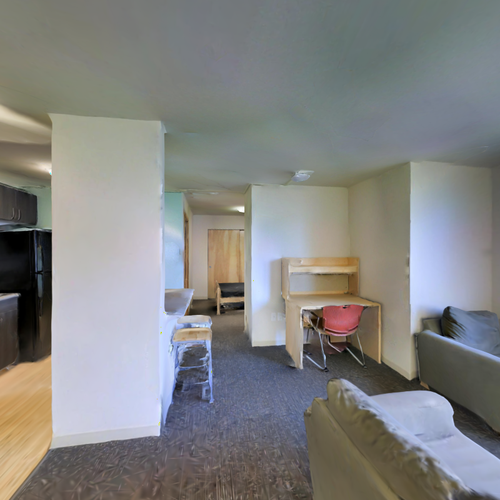} & \includegraphics[height=\rendersize]{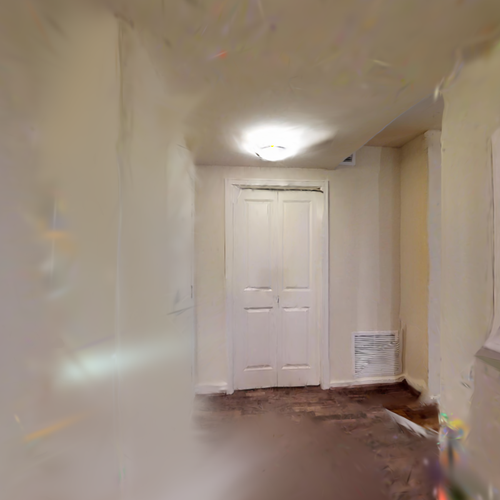} \\    
 \qquad \qquad UPEN & \includegraphics[height=\rendersize]{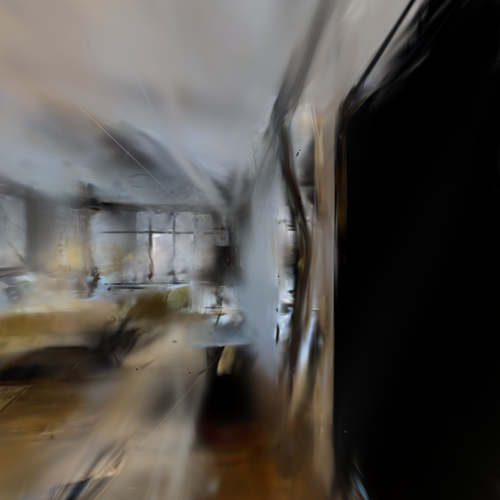} & \includegraphics[height=\rendersize]{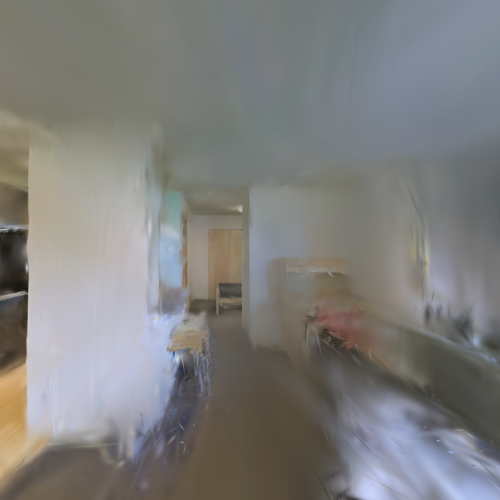} & \includegraphics[height=\rendersize]{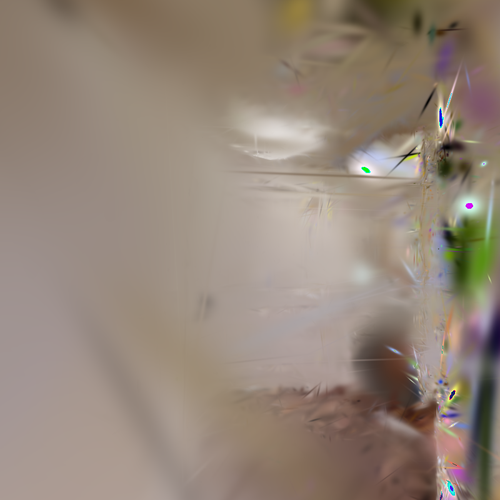} \\ 
\qquad \quad ExplORB & \includegraphics[height=\rendersize]{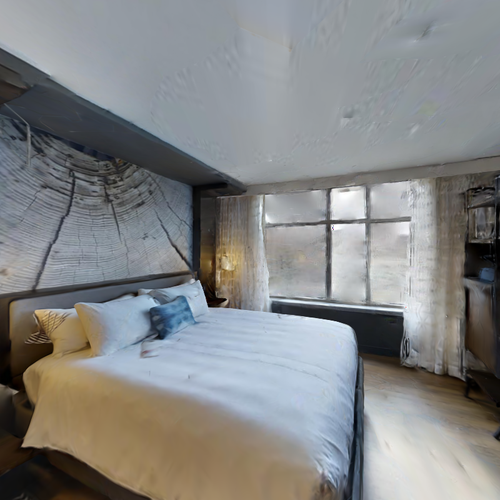} & \includegraphics[height=\rendersize]{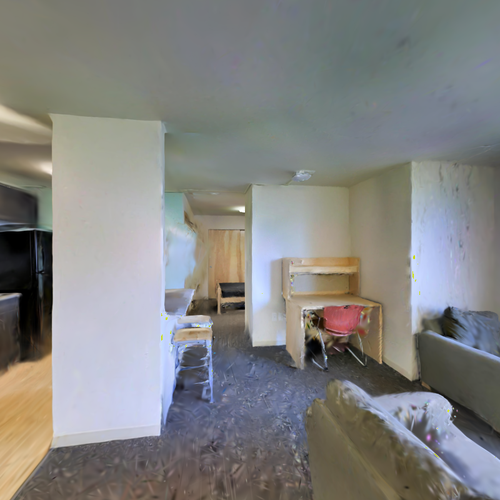} & \includegraphics[height=\rendersize]{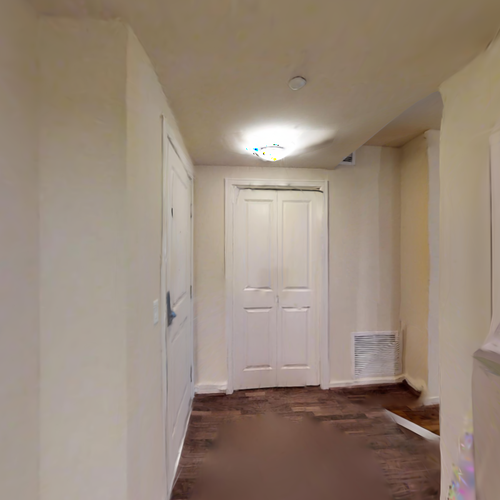} \\ 
 \qquad \qquad FBE & \includegraphics[height=\rendersize]{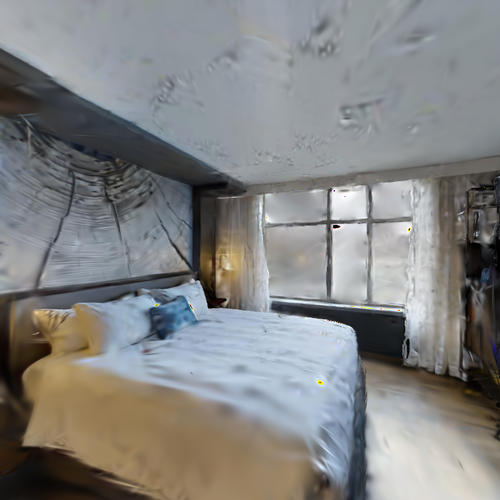} & \includegraphics[height=\rendersize]{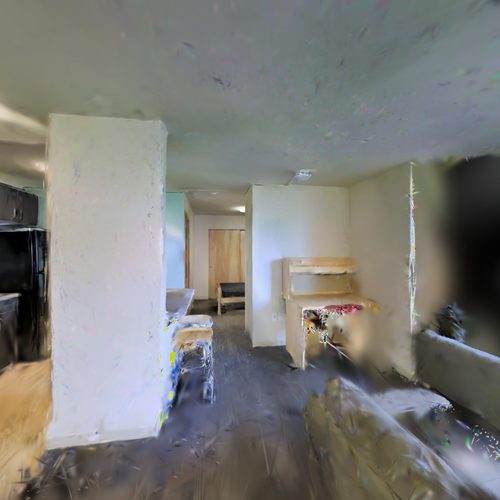} & \includegraphics[height=\rendersize]{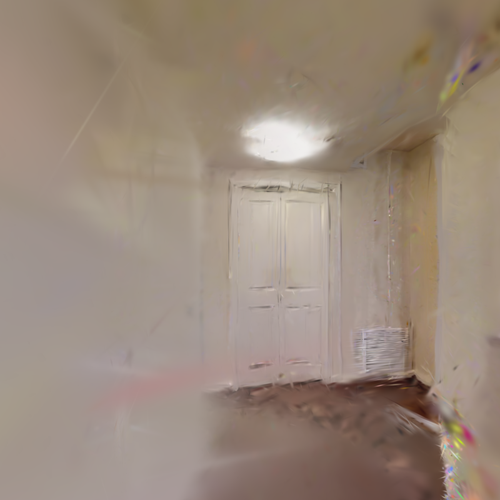} \\ 
 \qquad \quad Ours & \includegraphics[height=\rendersize]{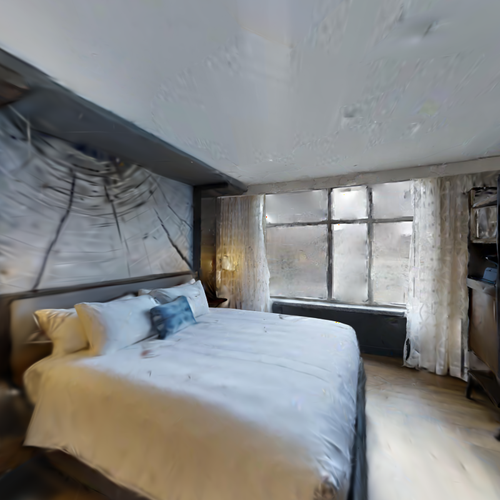} & \includegraphics[height=\rendersize]{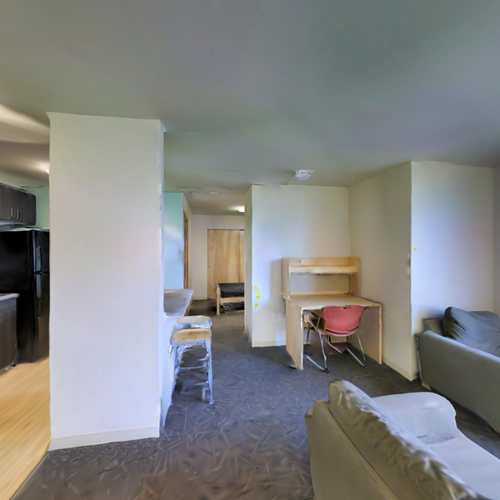} & \includegraphics[height=\rendersize]{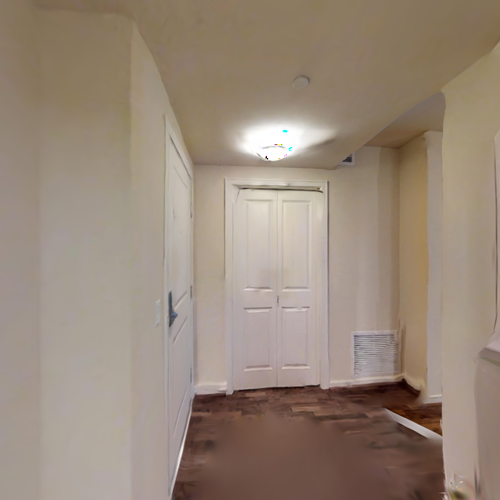} \\ 

\end{tabular}

\caption{{\bf Test Rendering Qualitative Visualization on Gibson Dataset} 
All the renderings are from the test view of the Gibson dataset.
}\label{fig:qual-render-gibson}
\end{figure*}

\begin{figure*}[!t]
\centering

\begin{tabular}
{>{\begin{sideways}}p{0.4in}<{\end{sideways}}p{0.25\textwidth}p{0.25\textwidth}p{0.25\textwidth}}   
 \qquad \qquad UPEN & \includegraphics[height=\rendersize]{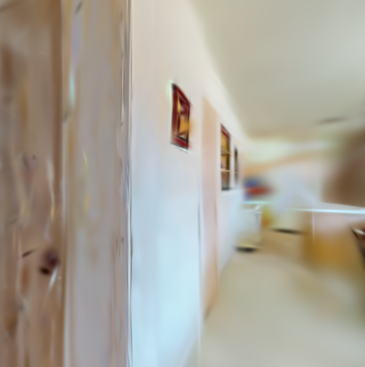} & \includegraphics[height=\rendersize]{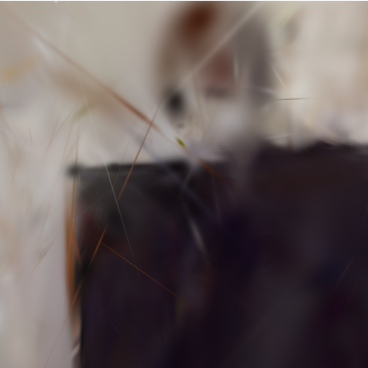} & \includegraphics[height=\rendersize]{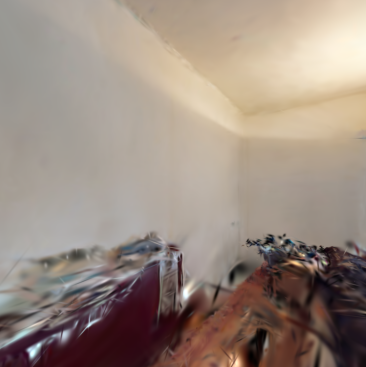} \\ 
 \qquad \qquad FBE & \includegraphics[height=\rendersize]{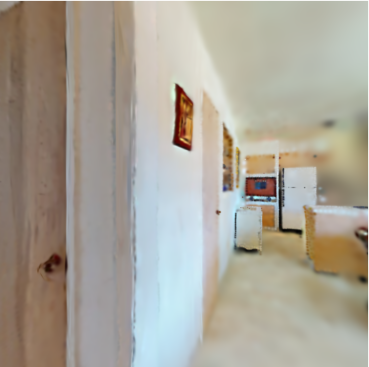} & \includegraphics[height=\rendersize]{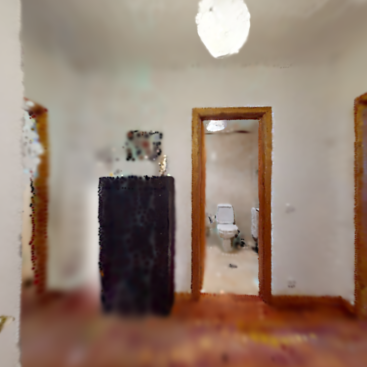} & \includegraphics[height=\rendersize]{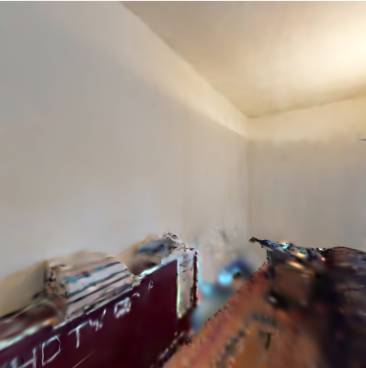} \\ 
 \qquad \quad Ours & \includegraphics[height=\rendersize]{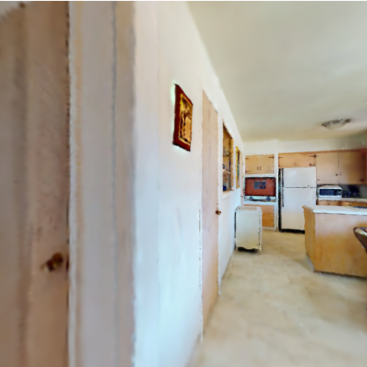} & \includegraphics[height=\rendersize]{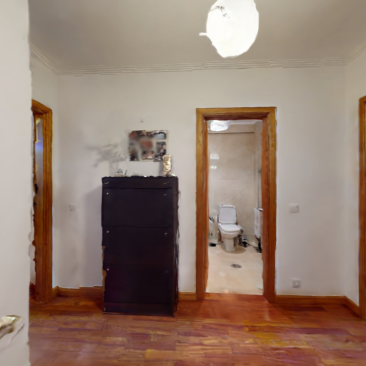} & \includegraphics[height=\rendersize]{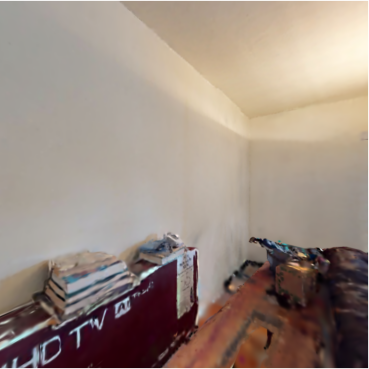} \\ 

\end{tabular}

\caption{{\bf Test Rendering Qualitative Visualization on Habitat-Matterport 3D Dataset} 
All the renderings are from the test view of the Habitat-Matterport 3D dataset.
}\label{fig:qual-render}
\end{figure*}

\end{document}